\documentclass{article}

\usepackage[margin=1in]{geometry}
\usepackage{mathpazo}
\usepackage{booktabs}

\usepackage[margin=1in]{geometry}
\usepackage{mathpazo}

\usepackage{booktabs}
\usepackage{amsmath,amssymb}
\usepackage{enumitem} %
\usepackage{amsmath,amssymb,amsthm}
\usepackage{xspace}
\usepackage{graphicx}
\usepackage{color,soul}
\usepackage[colorlinks=true,linkcolor=blue,citecolor=blue,urlcolor=blue]{hyperref}
\usepackage[labelfont=bf]{caption}
\usepackage{mathtools}
\usepackage[labelfont=bf]{subcaption}
\usepackage{comment}
\usepackage{algorithm,algpseudocode}
\usepackage[T1]{fontenc}
\usepackage[capitalise, noabbrev]{cleveref}

\usepackage[colorinlistoftodos,textsize=tiny,textwidth=2.1cm,shadow,loadshadowlibrary]{todonotes}
\newcommand{\dsdm}{{\sc DsDm}}

\usepackage[backend=biber,style=alphabetic,natbib=true,maxbibnames=99,minalphanames=3,maxcitenames=1]{biblatex}
\title{Small-to-Large Generalization: Data Influences Models Consistently Across Scale}
\author{
    Alaa Khaddaj \\
    \texttt{alaakh@mit.edu} \\
    MIT
    \and
    Logan Engstrom \\
    \texttt{engstrom@mit.edu} \\
    MIT
    \and
    Aleksander M\k{a}dry \\
    \texttt{madry@mit.edu} \\
    MIT
}
\date{}

\makeatletter
\let\c@figure\c@table
\makeatother
\begin{document}

    \setcounter{tocdepth}{3}

    \maketitle
    \begin{abstract}
        Choice of training data distribution greatly influences model behavior. Yet, in
large-scale settings, precisely characterizing \textit{how} changes in training
data affects predictions is often difficult due to model training costs. Current
practice is to instead extrapolate from scaled down, inexpensive-to-train proxy
models. However, changes in data do not influence smaller and larger models
identically. Therefore, understanding how choice of data affects large-scale
models raises the question: how does training data distribution influence model
behavior across compute scale? We find that small- and large-scale language
model predictions (generally) \textit{do} highly correlate across choice of
training data. Equipped with these findings, we characterize how proxy scale
affects effectiveness in two downstream proxy model applications: data
attribution and dataset selection.

    \end{abstract}

    \section{Introduction}
    \label{sec:intro}
    When training large-scale models, we often want to understand how changing the
training data distribution influences model behavior. For example, we may ask: does adding a
data source improve accuracy? Does removing a data source increase toxicity?
However, answering such questions is difficult in practice as the cost of model
training makes training on each data distribution (and comparing the resulting
models) infeasible.

To overcome compute costs, current practice is to approximate large-scale model
behavior with that of small-scale models. In this approach, one (a) calculates
how a given change in data distribution changes small-scale (low-cost) models
(e.g., by retraining small models with and without the change), then (b)
extrapolates the corresponding influence for large-scale model predictions using
insights from (a). Indeed, small-scale \textit{proxy models} are a standard
primitive in methods for dataset selection and
cleaning~\citep{engstrom2024dsdm,mosaicml2023introducing,xie2023doremi,chen2023skill}.

Nevertheless, there is yet no precise characterization of
when proxy models are effective. After all, model behavior often changes across
scale \citep{wei2022emergent}; thus, changes in data may not influence small-
and large-scale models identically. Understanding how training data changes
large-scale model behavior therefore hinges on the question: how does training
data influence model behavior across compute scale?

\begin{figure}[h]

    \centering
    \includegraphics[width=\linewidth]{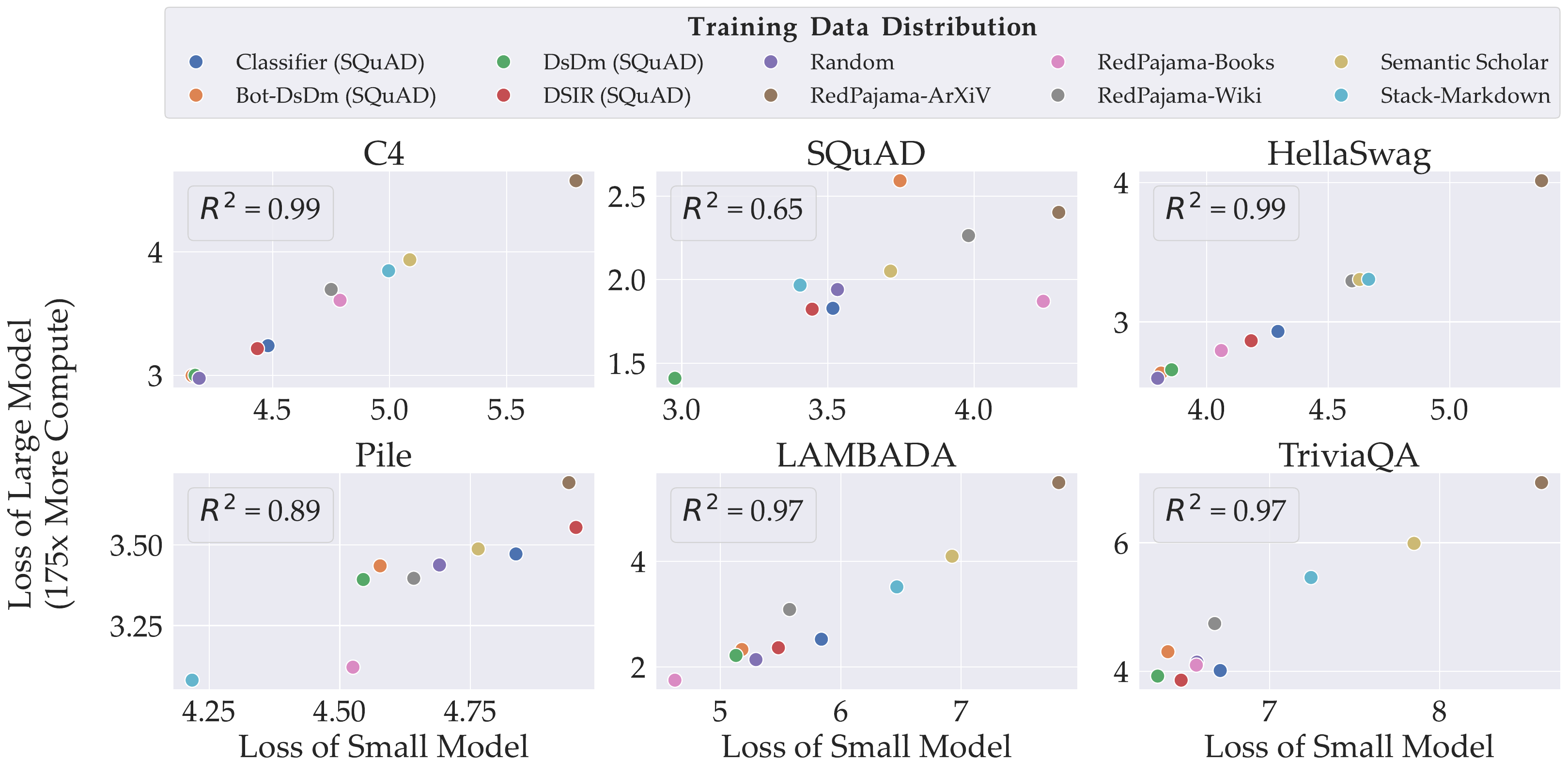}
    \caption{Proxy-model test loss highly correlates with large-model test loss
    across choice of training data distribution, even across a large gap in
    scale. Above, we plot the losses of a small-scale proxy (57M parameters)
    compared to that of the reference model (760M parameters). Here, the small
    scale model trains with 175\texttimes{} less compare than the reference
    model. Each column represents model loss on a different test distribution,
    ranging from LM benchmarks (SQuAD/HellaSwag) to pretraining data
    distributions (the Pile).}
    \label{fig:small_large_losses_nlp}
\end{figure}

\paragraph{Contributions.}
After training language models (LMs) on a diverse set of training data
distributions at different scales, we find that the answer is nuanced. On one
hand, choice of training data distribution generally affects model predictions
(very) similarly along compute scale (down to 175\texttimes{} smaller than the large-scale
reference model, cf. \cref{fig:small_large_losses_nlp}). Indeed, such a
relationship even holds when proxy models are so small that their predictions
are as accurate as \textit{randomly guessing}.

On the other hand, however, our results also indicate that proxy models are not
a panacea: we identify setups for which proxy model predictions do not correlate
well with larger models. We find that only (very) small proxy models---those
370\texttimes{} smaller than the large-model class of interest---tend to predict
larger-scale model behavior poorly.

Equipped with these findings, we then characterize the relationship between
proxy model scale and performance in two downstream proxy model applications:
data attribution (in vision settings) and dataset selection (in an LM setting)
for large models. In both applications, we find that orders-of-magnitude smaller
proxy-models can be as effective as using the original, larger-scale model of
interest directly---but also that there is a clear trade-off between performance
and proxy-model size at the smallest scales we study.

    \section{Data Influence Across Scale}
    \label{sec:intuition}

We seek to characterize how choice of training data influences model behavior
across compute scale (i.e., the amount of compute used to train a model). To do
so, we compare how changes in training data distribution affect large-scale
model predictions compared to those of small-scale proxy models trained on the
same data distributions. Correlating these differences across a diverse set of
training data distributions, we find that training data generally influences
model predictions similarly across scale---but that the degree of correlation
depends on both the exact choice of test distribution and proxy model scale. In
what follows, we first describe our experimental setup, then detail results (see
\cref{sec:app_sim} for additional details).

\subsection{Setup}
We study how changes in data distribution affect the behavior of small {\it proxy}
models compared to the behavior of a larger \textit{reference} model class. We
select 760M parameter language models as the reference model class (the largest
setting that we can study in our available, academic-level compute budget). Our
proxy models range in size from 40M parameters to 760M parameters, with each
model training on a number of tokens determined by Chinchilla-optimal
token-to-parameter ratios~\citep{kaplan2020scaling}. In relative terms, these
model train with down to 370\texttimes{} less compute than the reference model
despite only having (at most) 19\texttimes{} fewer parameters (as they are
trained with chinchilla-optimal token-to-parameter ratios).

We measure how model behavior changes across 10 separate training distributions:
6 {\it data-sources} (i.e., sampled from a single data source like
Wikipedia~\citep{wikimedia2023wikimedia}) and 4 {\it selection-induced}
distributions (i.e., data selected with one of three dataset selection methods:
{\sc DsDm} \citep{engstrom2024dsdm}, DSIR \citep{xie2023data} and
Classifier-based approach \citep{brown2020language} using various target tasks).
After training (separate) models on each of these training datasets, we compare
the resulting model behavior (losses) on 6 test datasets: C4
\citep{raffel2020exploring}, the Pile \citep{gao2020pile}, SQuAD
\citep{rajpurkar2016squad}, LAMBADA \citep{paperno2016lambada}, HellaSwag
\citep{zellers2018swag} and TriviaQA \citep{joshi2017triviaqa}.

\subsection{Results}
At a high level, we find that changes in training data distribution (generally)
affect small- and large-scale model predictions similarly---even when the small
proxy model is trained with much less compute than the large reference
model in relative terms. We use the following basic primitive to study the effect of
training data distribution: given downstream task, we measure
the correlation of small- and large-scale losses across training data distributions. To
obtain these results, we train small- and large-scale models on \textit{each}
training data distribution (one for each scale of model), and record the
empirical loss of each of these models on downstream tasks.

We begin by studying the behavior of a single proxy model scale: 57M parameter
proxy models. We relate in \cref{fig:small_large_losses_nlp} the losses of 57M parameter proxy
models to those of the reference model class across different training data
distributions, while varying (in each panel) the choice of downstream task.
These proxy model losses (generally) highly correlate with those of large-scale
models across training dataset, implying that choice of training dataset
similarly changes both 57M and large (760M) model predictions---despite the
proxy models training with 175\texttimes{} less compute.

\begin{figure}[h]
    \centering
    \includegraphics[width=0.45\linewidth]{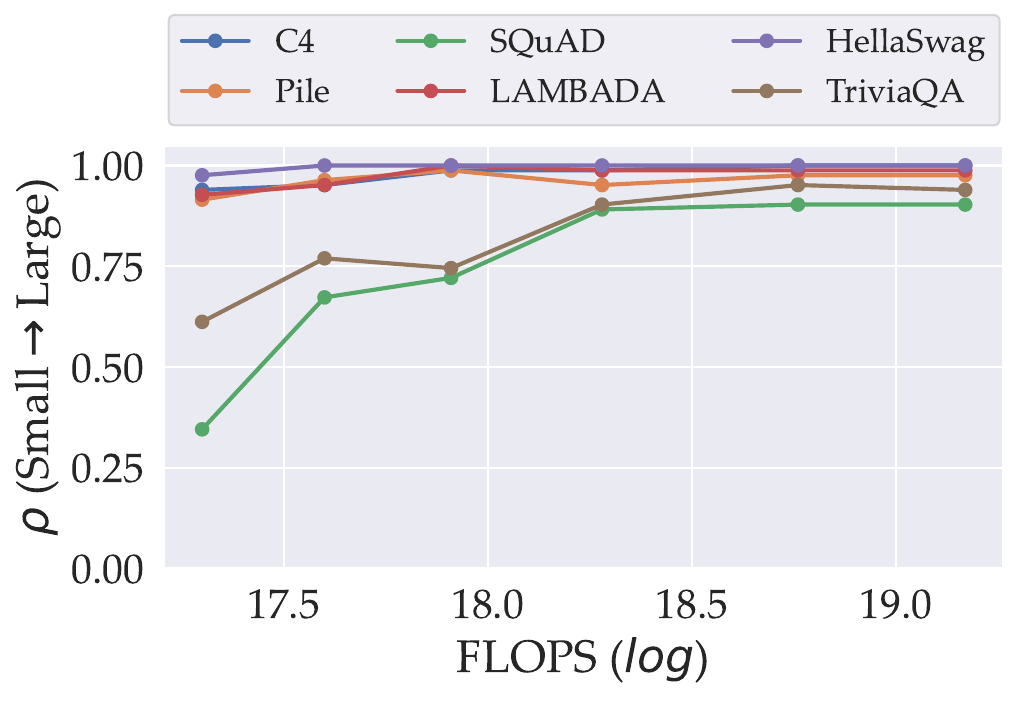}
    \caption{Correlation between large- and small-scale model predictions is
    consistently high, even across large gaps (orders of magnitude) in training
    compute scale. We plot small- to large-scale correlation against small-scale
    proxy model compute. There is also large variation across choice of test
    set: correlation is consistently high on four of six tasks, while losses on
    SQuAD and TriviaQA correlate less.}
    \label{fig:r2_vs_compute}
\end{figure}

To further study the role of proxy model scale, we relate in \cref{fig:r2_vs_compute}
proxy model scale with the correlation between proxy and reference model
predictions. We find that, as in the case of the 57M proxy model, losses are
highly correlated. In general, losses are more correlated for proxy
models that are closer in scale to the reference model.

However, our results also indicate that proxy models are not \textit{always} reliable:
the correlation between reference and proxy model predictions is highly
dependent on (a) the gap in scale between the proxy and reference models (much
smaller proxies are more mismatched) and (b) the exact choice of downstream task
(proxy predictions are less correlated with reference model predictions on
specific test distributions). For example, consider the smallest proxy model in
\cref{fig:r2_vs_compute} (40M models, which use 370\texttimes{} less compute than the
large model of interest). This class of model is highly correlated
with the reference model on all the downstream tasks except two: SQuAD and
TriviaQA (cf. \cref{fig:r2_vs_compute} for a detailed view).

\subsection{Intriguing properties of proxy models}
We observe two additional properties of the relationship between proxy and
reference models.

\begin{figure}[h]
    \centering
    \includegraphics[width=0.5\linewidth]{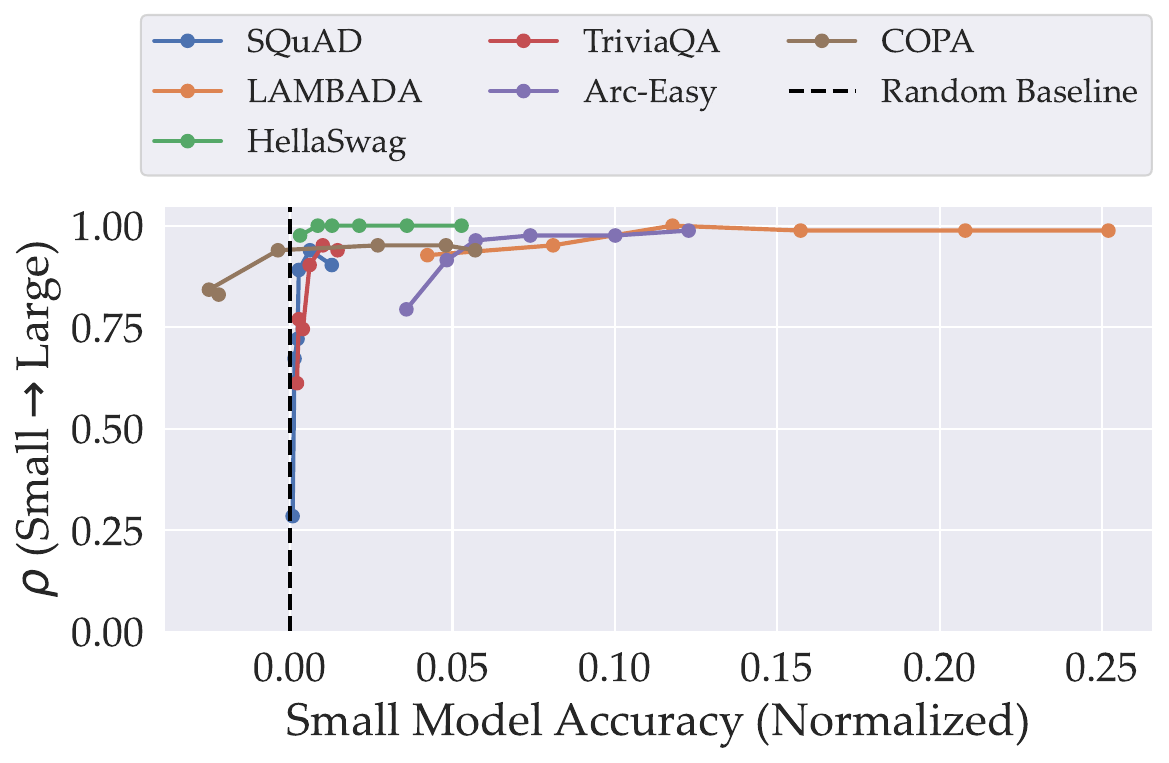}
    \caption{Proxy models can be highly predictive of large-scale model
    predictions even when predicting as well as randomly on a given test set. We
    plot small- to large-scale loss correlation against small-scale proxy model
    accuracy on the given task, normalized to show improvement over outputting a
    random guess (in absolute accuracy). On a number of test sets, proxy models
    perform no better than random guessing, but still highly correlate with the
    reference model (which always achieves significantly better than random
    guessing).}
    \label{fig:corr_vs_small_acc}
\end{figure}

\paragraph{Proxy models are effective regardless of accuracy.}
We find that proxy model predictions for a given task can highly correlate with
those of large-scale reference models {\it even} when the proxy models predict near the level of
\textit{random guessing} on that task. Indeed, relating proxy model accuracy
against correlation with reference model predictions in \cref{fig:corr_vs_small_acc},
we find that in two tasks---HellaSwag and COPA---small-scale proxy models
achieve random-guessing level (or worse) accuracy while still highly correlating
with large-scale models.

\begin{figure}[h]
    \centering
    \includegraphics[width=\linewidth]{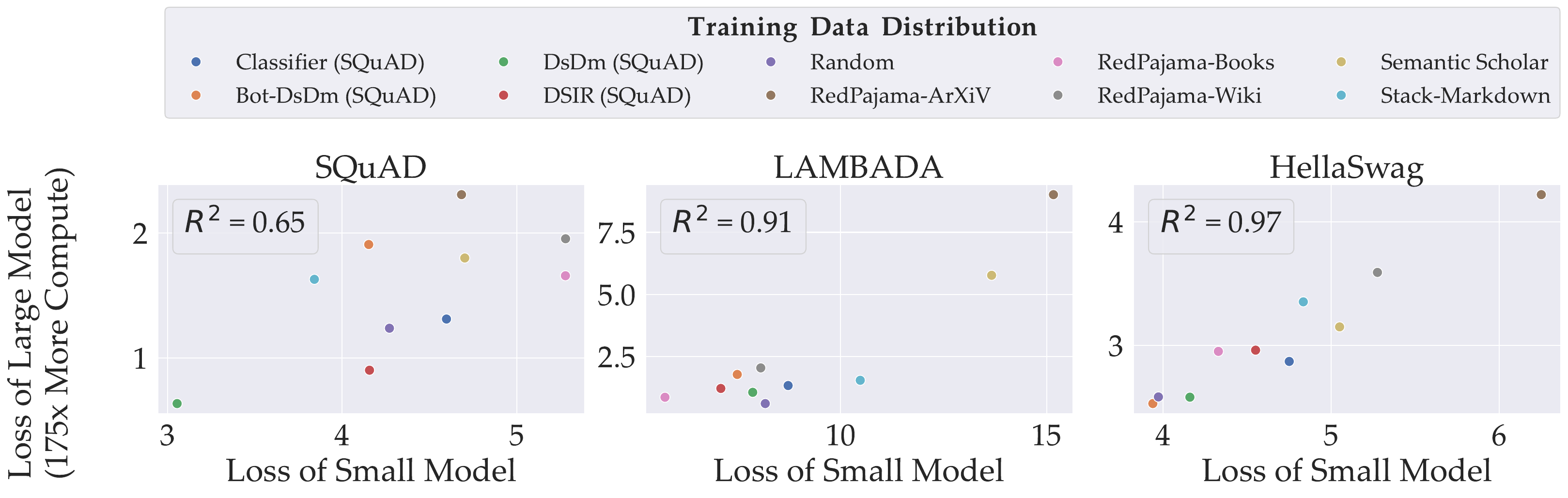}
    \caption{Proxy model predictions can highly correlate with those of the
        reference model on \textit{individual} test samples. We visualize loss
        on individual samples for each scale model across varying training
        datasets. The proxy model here is 57M parameters, training with around
        175\texttimes{} the compute of the 760M reference model. See a
        distributional plot (showing the correlation across \textit{all} samples
        on each test set) in Figure~\ref{fig:small_large_r2_dist}.}
    \label{fig:small_large_losses_seq_57m}
\end{figure}

\begin{figure}[h]

    \centering
    \includegraphics[width=\linewidth]{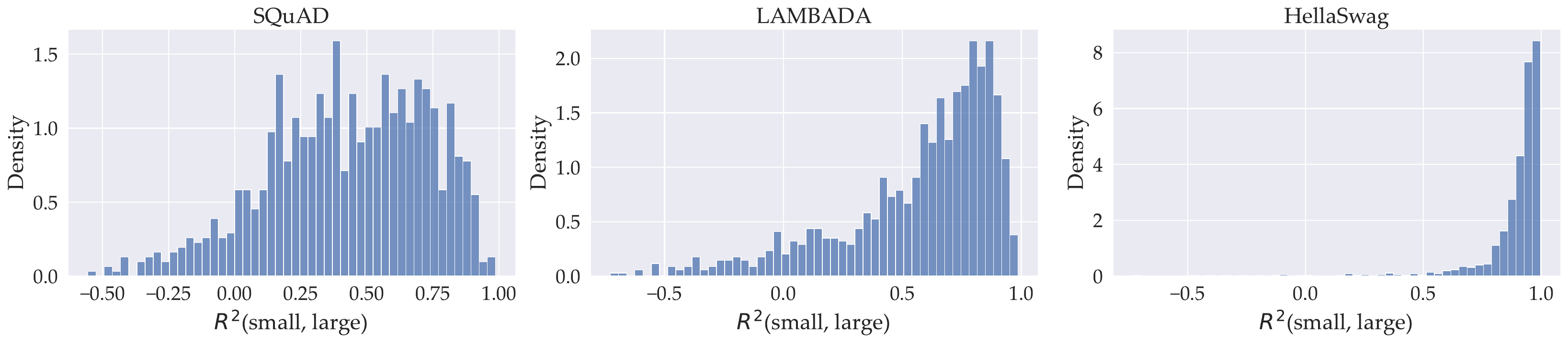}

    \caption{The correlation between large- and small-scale model losses on
    individual samples is highly dependent on the test distribution. We show a
    histogram of the correlation between large model and proxy model predictions
    on individual test samples for the test distribution in each column. We plot
    the coefficient of determination ($R^2$) between the losses of the small and
    large models on all examples in the downstream task.}
    \label{fig:small_large_r2_dist}

\end{figure}

\paragraph{Proxy models are (often) effective at a per-sample level.}
We have thus far only studied the relationship between \textit{average} losses
achieved by proxy and reference models on each test task. To better characterize
when proxy models match the reference model, we inspect similarity between
small- and large-scale model predictions on \textit{individual samples}---for
individual test samples---in \cref{fig:small_large_losses_seq_57m}. Our results
indicate that proxy model predictions on individual samples can highly correlate
with those of large models, depending on the choice of test dataset. On a
population view, however, the picture is more nuanced: while proxy model
predictions highly correlate with reference model predictions on the great
majority of HellaSwag samples, they do not correlate as well on SQuAD samples
(cf. \cref{fig:small_large_r2_dist}).

    \section{Proxy Models in Downstream Applications}
    \label{sec:methodology}
    Proxy model predictions generally highly correlate with reference model
predictions across training distribution choice. However, at small proxy model
scales this relationship can break down, suggesting that there is a fundamental
trade-off between proxy compute scale and effectiveness.
To understand how the proxy scale affects the utility of proxy models in
downstream tasks, we characterize the role of proxy model scale in two
downstream applications: attributing training data and selecting training data.

\subsection{Attributing Training Data with Proxy Models}
Data attribution methods analyze model behavior in terms of the training data \citep{koh2017understanding,ilyas2022datamodels}.
While these methods are helpful in tasks like dataset selection \citep{engstrom2024dsdm}
and model debugging \citep{ilyas2022datamodels}, they also tend to require compute that scales
with the model size and the training dataset size. This requirement often
makes data attribution prohibitively expensive in large-scale settings \citep{koh2017understanding,schioppa2022scaling,grosse2023studying}.
To make data attribution feasible at this scale, common practice is to instead
attribute for a smaller proxy model, then use the result to attribute for the
original model of interest \citep{engstrom2024dsdm}.

\subsubsection{Preliminaries}
\label{sec:prelims}
We start by defining data attribution within the datamodeling
framework~\citep{ilyas2022datamodels}. Consider a training dataset $S = \{(x_1,
y_1), \ldots, (x_n, y_n)\}$ of $n$ input-label pairs, and let $\theta(D)$ be the
parameters of a classifier trained on subset $D$ of $S$. Then, given a
sample $z = (x, y)$, let $f(z; \theta(D))$ be the loss of the classifier on $z$
after training on subset $D$ of the training set.

A \textit{datamodel} for heldout sample $z$ is a simple (learned) function that estimates
the final model loss on $z$ as a function of the subset $D$ used to train the
model. For convenience, this is the function
$$
\hat{f}_z(D) \approx f(z; \theta(D)),
$$
which maps choice of training dataset to the loss of the resulting model on $z$.
Intuitively, a datamodel $\hat{f}_z$ should accurately predict model loss after
training on any given train subset $D$.

Previous work has found that the loss $f(z; \theta(D))$ can be approximated by
\textit{linear} datamodels, or datamodels that parameterize each training datapoint as
contributing a fixed amount to the loss when included in the training dataset.
That is, we can approximate the model loss reliably using the linear datamodel
$\hat{f}_z$ parameterized as:
\begin{equation}
    \hat{f}_z(D) \coloneqq \sum_{i\in D} \tau(z)_i,
    \label{eq:dm_loss}
\end{equation}
where $\tau(z)_i$ is a weight representing the ``importance'' of training
example $i$ on predicting the heldout sample $z$ correctly.

\paragraph{Estimating datamodel weights.}
Families of approaches for estimating datamodel weights range from influence
functions~\citep{koh2017understanding,grosse2023studying} to resampling
estimators~\citep{feldman2019does,ilyas2022datamodels}. In this work, we
estimate datamodels using an influence function-based method called
TRAK~\citep{park2023trak}. Briefly: TRAK estimates datamodel weights by (a)
linearizing (trained) model output with respect to the model weights and then
(b) calculating influences for this linearization~\citep{koh2017understanding}.
See \cref{sec:app_background} for full details and setup.

\paragraph{Evaluating datamodels.} We evaluate datamodels with the Linear
Datamodeling Score, or LDS~\citep{ilyas2022datamodels,park2023trak}, a standard
approach for evaluating data attribution methods
~\citep{Bae2024-qo,zheng2023intriguing,choe2024dataworthgptllmscale,lin2024efficient,
georgiev2023journey, deng2024ensembleattrib}. For a heldout sample
$z$, LDS measures the correlation between datamodel prediction of model loss and
the actual model loss across $m$ randomly sampled training subsets $D_i$ (e.g.,
a common choice is to randomly choose fixed-size subsets of the training set).
Specifically, the LDS for our linear datamodels is exactly the (Spearman)
correlation:
\begin{equation}
LDS(\tau(z), z) := \rho_\mathrm{spearman} \big(\underbrace{\vphantom{\sum\nolimits_{k \in D_k}}f(z; \theta(D_i)) : i \in [m]}_{\text{actual model loss}}, \quad \underbrace{\sum\nolimits_{k \in D_i} \tau(z)_k : i \in [m]}_{\text{datamodel-predicted loss}} \big).
\end{equation}
Intuitively, a datamodel that perfectly captures model loss would have an LDS of $1$, and a
datamodel that does not correlate with the model loss would have an LDS of $0$.
In this
work, we measure the expected LDS over a given test distribution (by averaging
LDS over test samples).

\subsubsection{Experimental Results}
We study how well datamodels computed from smaller proxy models approximate the
actual loss of the reference model in two supervised computer vision settings:
ImageNet-1k~\citep{russakovsky2015imagenet} and
CIFAR-10~\citep{krizhevsky2009learning}.

\paragraph{Setup.} We estimate datamodels for ResNets \citep{he2015residual}
across a variety of model widths (ImageNet: the largest model class has a width
$10^4$ times larger than the smallest; in CIFAR-10 this relative range is
$10^5$). We then evaluate these datamodels by measuring the LDS with respect to
the predictions of the \textit{largest model class} (a $10^8$ parameter ResNet
for ImageNet and $10^9$ for CIFAR-10). For additional details and results, see
\cref{sec:app_vision_setup}.

\begin{figure}[t]

    \begin{subfigure}[c]{0.48\linewidth}
        \centering
        \includegraphics[width=\linewidth]{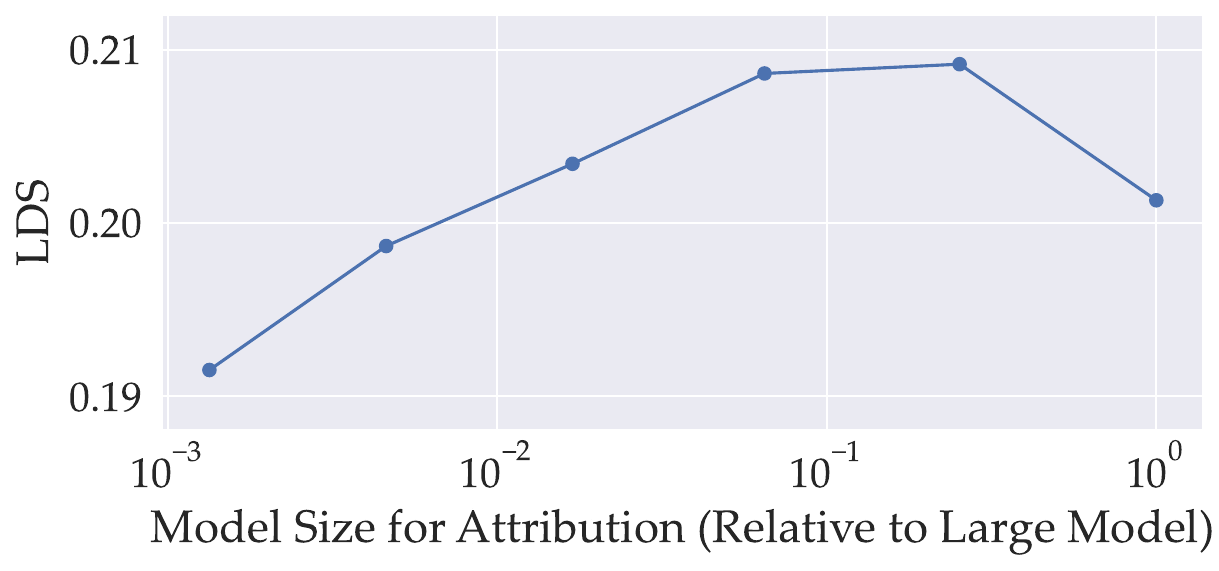}
        \caption{ImageNet}
        \label{fig:imagenet_corr_vs_flops}
    \end{subfigure} \hfill
    \begin{subfigure}[c]{0.48\linewidth}
        \centering
        \includegraphics[width=\linewidth]{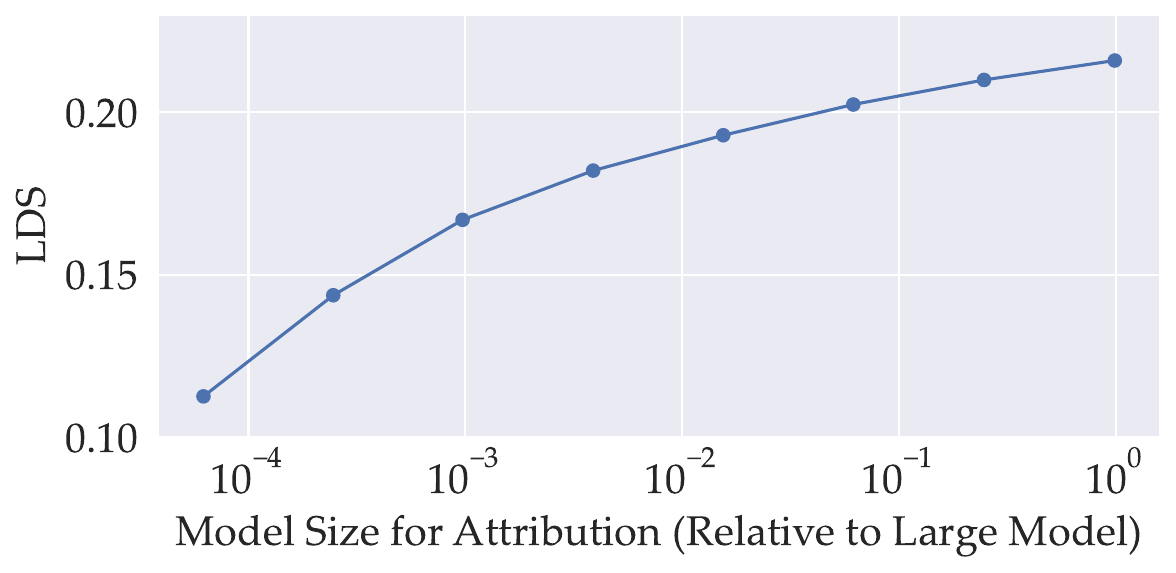}
        \caption{CIFAR-10}
        \label{fig:cifar_corr_vs_flops}
    \end{subfigure} %

    \caption{
        In both plots, the $x$-axis represents the amount of compute required to get the attribution scores of a given model, compared to the large model, and the $y$-axis represents how well the attribution scores of a given model size can predict the output of the largest model on
        {\bf (a)} CIFAR-10 and {\bf (b)} CIFAR-100 respectively \citep{krizhevsky2009learning} (see \cref{sec:methodology} for details on the metric).
    }
    \label{fig:corr_vs_flops}

\end{figure}

\paragraph{Results.}
Small proxy models yield datamodel estimates that are similar in
effectiveness to those calculated with the actual, large-scale model reference model.
Relating proxy model size to LDS in \cref{fig:corr_vs_flops} (left) in
the ImageNet setting, we find that LDS decreases in relative terms by (at most)
$10\%$ (from $0.21$ to $0.19$) across \textit{all} proxy models, even those
that are 1,000\texttimes{} smaller than the reference model. In the
CIFAR-10 setting (cf. \cref{fig:corr_vs_flops} right), the LDS only greatly degrades
after proxy models are more than 1,000\texttimes{} smaller than the reference model.

\begin{figure}[t]

    \centering
    \includegraphics[width=\linewidth]{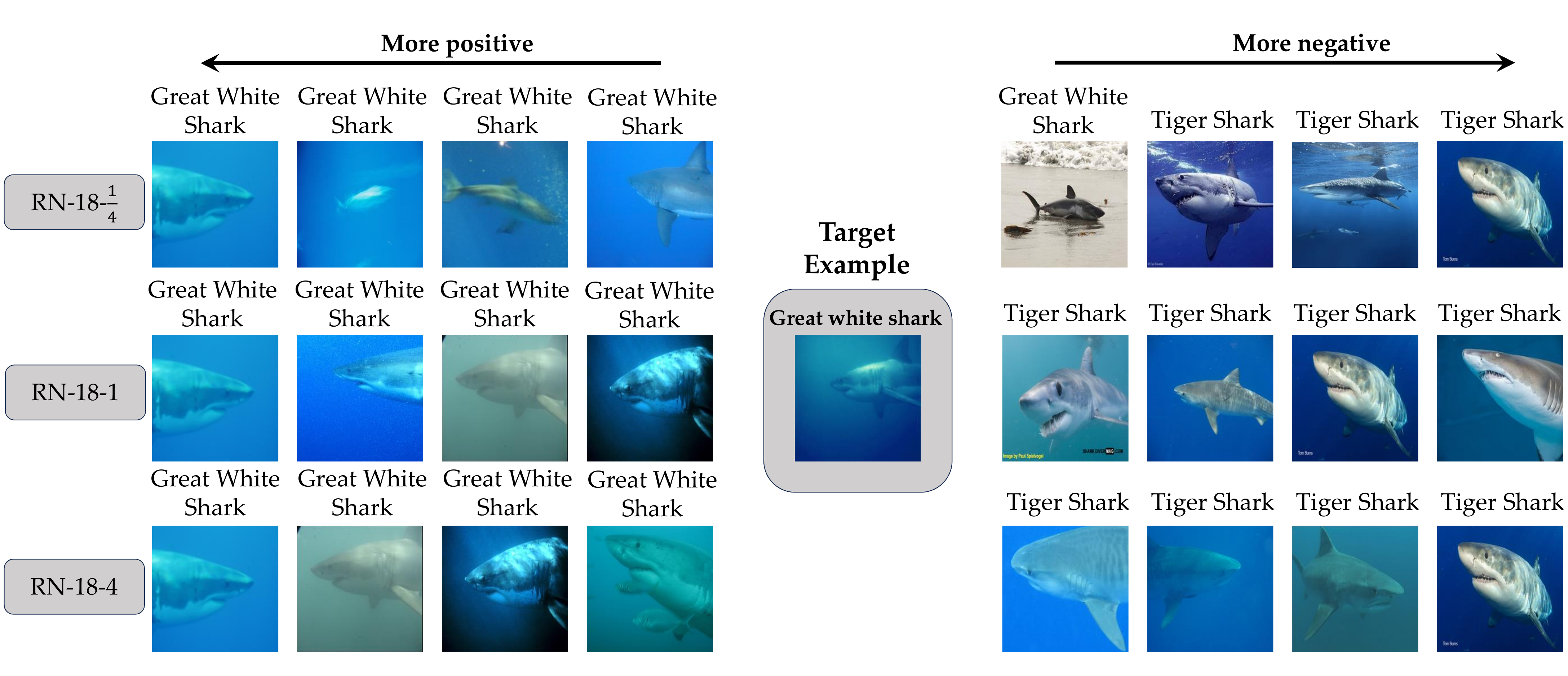}

    \caption{Most helpful ({\it left}) and most detrimental ({\it right})
    examples for the outputs of models of different sizes are similar. The most
    helpful and most detrimental examples for the given target example (center)
    are shown according to each model size (row). We observe a large overlap
    between these examples. More examples in \cref{app:results_qual_sim_vis}.}
    \label{fig:top_bottom_visual}

\end{figure}

More qualitatively, we also compare the ``top'' and ``bottom'' training examples
(by datamodel weight) for a given test sample in
Figure~\ref{fig:top_bottom_visual} across proxy model sizes. Intuitively, these
examples are the ones that (according to the datamodels and by linearity) most
improve and most hurt, respectively, model performance if included in the
training set. We find that, qualitatively, these top and bottom examples
generally overlap across model scales and often have visually similar
attributes. See more examples in \cref{sec:app_results}.

\paragraph{Limitations.}
We note that all the measured LDS correlations are seemingly small. The peak LDS
measured in this work is roughly $0.21$ for ImageNet, which indicates that we
cannot exactly predict model outputs for a given training set. These LDS numbers
are primarily due to (a)
limitations in current datamodel estimation methods (e.g., state-of-the-art
methods achieve similar LDS for CIFAR-10~\citep{Bae2024-qo}) and (b) inherent
randomness during training\footnote{Computing LDS requires retraining models on
different subsets, and the inherent randomness involved in retraining models
results in an irreducible error.}. The room for improvement indicates that it is
possible that future, more effective datamodel estimators will behave
qualitatively differently from current estimators---and that the precise
trade-off between model scale and datamodel quality could change as well.

    \subsection{Selecting Training Data with Proxy Models}
In dataset selection, the goal is to choose the best possible
training dataset out of a larger pool of candidate data.
In this work we focus on model-aware dataset selection methods, which
use the learning algorithm to select data
\citep{xie2023doremi,engstrom2024dsdm,xia2024less}.
Consequently, the compute cost of these methods typically grows with
the cost of the learning algorithm itself\footnote{In comparison, model-free dataset
selection methods clean data without considering the model, instead using e.g.,
heuristics that capture intuitive notions of data quality \citep{li2024datacomplm}.}.
As a result, model-aware dataset selection often leverage smaller proxy models
for selection in place of the original (more expensive) model.
In this section, we
characterize the relationship between dataset selection effectiveness and proxy
model size.

\subsubsection{Preliminaries}
Following previous work, we formalize data selection as the problem of finding
the subset of data, out of a larger pool of candidate data, that maximizes
downstream trained model accuracy on a given
task~\citep{engstrom2024dsdm,xia2024less}. Here, selecting training data is
a supervised learning task: given (maybe only a few) samples from the
test distribution, choose the data that maximizes trained model performance.
In this work, we select training data with \dsdm{}~\citep{engstrom2024dsdm}, a
method that uses datamodels to select data \citep{ilyas2022datamodels}.
We refer the reader to \cref{sec:app_dsdm} for more details
on \dsdm{}.

A major challenge with this approach is the compute required to calculate
the datamodels for langugage models with even as few as 1B parameters. To reduce the compute
cost, \citet{engstrom2024dsdm} computed the datamodels for a smaller proxy model
and used these datamodels to select the training subset. We explore in this section
the tradeoff between the scale of the proxy model to attribute and the performance
of the large reference model trained on the training subset selected using
the datamodels of the proxy model.

\subsubsection{Experimental Results}
We study how the size of the small proxy model used for dataset selection
affects model performance on two downstream tasks, SQuAD and LAMBADA.

\paragraph{Setup.}
We consider a language modeling (LM) setting where GPT-2 style
LMs \citep{radford2019language} are pretrained on subsets of the
MPT dataset \citep{mosaicml2023introducing}\footnote{Our subset of the MPT dataset
contains 160B tokens.}
and evaluated on two popular zero/few-shot classification tasks:
SQuAD \citep{rajpurkar2016squad} and LAMBADA \citep{paperno2016lambada}.

Our large reference model is a 760M parameter LM\footnote{This model is the
largest we can study in our available, academic-level compute budget.}, and our
proxy model sizes range from 125M parameters to 760M parameters. We train all
models on datasets sized according to Chinchilla-optimal token-to-parameter
ratio \citep{hoffmann2022training}\footnote{We use the {\it llm-foundry}
repository \citep{mosaicml2023llm} for training and evaluating our models.}. We
calculate the datamodels for each of our proxy models, then select a subset of
the training dataset (using {\sc DsDm} \citep{engstrom2024dsdm}) to pretrain the
760M parameter reference model. As selection baselines, we consider reference
models trained on randomly-selected subsets with size dictated by the
Chinchilla-optimal token-to-parameter ratio. More details are included in
\cref{sec:app_lang_setup}.

\begin{figure}[t]

    \centering
    \includegraphics[width=\linewidth]{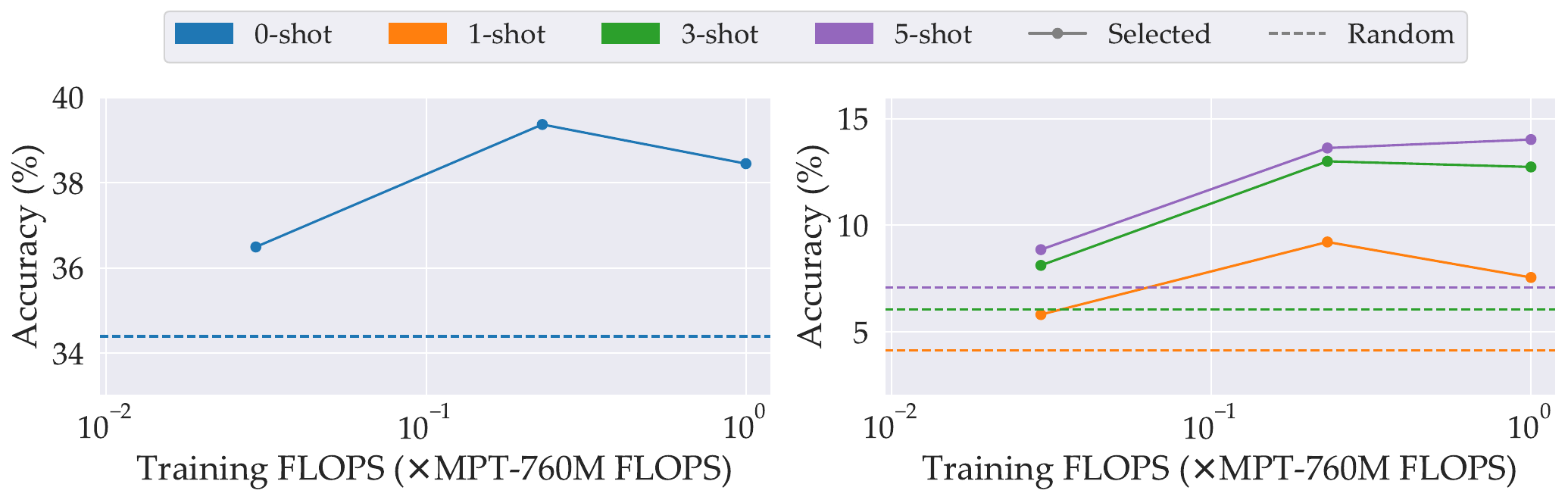}

    \begin{subfigure}[c]{0.48\linewidth}
        \centering
        \caption{LAMBADA}
        \label{fig:lambada_acc_vs_flops}
    \end{subfigure} \hfill
    \begin{subfigure}[c]{0.5\linewidth}
        \centering
        \caption{SQuAD}
        \label{fig:squad_acc_vs_flops_2}
    \end{subfigure}

    \caption{
        In both plots, the $x$-axis represents the amount of compute required to train a given proxy model (relative to training the large model) and the $y$-axis represents the accuracy on {\bf (a)} LAMBADA \citep{paperno2016lambada} and {\bf (b)} SQuAD \citep{rajpurkar2016squad} of a large model trained on a subset of the MPT dataset \citep{mosaicml2023introducing} selected using the attribution scores of a smaller model. The dashed line corresponds to the accuracy of a large model trained on a random subset of the same size as the selected dataset. Note that the training cost is only a fraction of the total attribution cost; see \cref{sec:app_compute}.
    }
    \label{fig:data_cleaning}

\end{figure}

\paragraph{Results.}
Models trained on data selected with {\sc DsDm} greatly improves over those
trained on randomly selected data, regardless of proxy model size (see
\cref{fig:data_cleaning}). We find that this improvement in downstream
performance does not drop until the proxy model training scale reduces to
4x less compute than the reference model. Our results
indicate that smaller proxy models mimic the behavior of reference models
enough to effectively select data, while simultaneously reducing the
compute cost.

    \section{Related Work}
    \label{sec:related}
    \paragraph{Using smaller proxy models.}
Small-scale \textit{proxy models} are a standard building block in approaches
that require understanding the role of data in large-scale models.
Proxy models are used to select and clean data~\citep{xie2023doremi,engstrom2024dsdm, chen2023skill, yu2024mates, li2024datacomplm}.
At a high level, these approachs train small-scale proxy models on candidate
data distributions, then analyze the resulting behavior to select the training
data for the large-scale models.

\paragraph{Data attribution.}
Data attribution has received increased interest lately. We discuss a few of these approaches in this section. For an extensive survey of prior work, we refer the reader to \citep{hammoudeh2022training}.
Some of the earliest approaches proposed the use of {\it influence functions} to approximate the effect of removing data points from the training dataset on a given parameter, without re-estimating the parameter \citep{hampel2011robust, koh2017understanding}. \citet{feldman2020what,ilyas2022datamodels} propose instead estimating empirically the effect of training data points on the model output by training several models on different subsets of the data and observing how the model output changes. Few other works have proposed different approaches to estimating these influences such as using Shapley values \citep{ghorbani2019data,jia2019towards,wang2021unified,shapley1951notes}, gradient-based approaches \citep{park2023trak,pruthi2020estimating} or representational similarity \citep{yeh2018representer,charpiat2019input}.

\paragraph{Similarities between models trained on the same dataset.}
A recent line of work argued that the data has a strong role in shaping the
behavior of the trained models. \citet{li15convergent} measured the extent to
which multiple networks learn the same set of features, while
\citet{hermann2020shapes} studied how different models learn easy and hard
features from a given dataset. \citet{nguyen2020wide} on the other hand focused
on how increasing the width of a network affects the learned representations.
More recently, \citet{vyas2023featurelearning} investigated how increasing the
width changes the properties of a model and its predictions at the example
level.

\paragraph{Relation between model behavior and size.}
Recent work argued that the behavior of large models is predictable from smaller
models under certain conditions \citep{yang2020tp4,yang2023tp6}. Specifically,
\citet{yang2020tp4} propose a parameterization of models, called $\mu P$ that
guarantees the output of a model converges as its size increases. $\mu P$ has
been very useful in practical setups, especially in ensuring good
hyperparameters found using small models can be transferred to large models
\citep{yang2022tp5}. Another work has argued that ``emergent'' abilities of
large models are a mirage \citep{schaeffer2023emergent} and that the reason
behind the emergence can be attributed to using {\it hard} metrics to measure
emergence (e.g., accuracy) rather than softer metrics (e.g., loss).

    \section{Conclusion}
    \label{sec:conclusion}
    In this work, we argue that the the choice of training data distribution
generally affects models across scale similarly, even when the difference in
compute is large (175\texttimes{} in our experiments). This trend, however, does
not always hold. In particular, given a large reference model and a {\it much}
smaller proxy model, we identify settings where the proxy model predictions do
not correlate well with the predictions of the reference model.
We then study the role of proxy model size in two
downstream applications: data attribution (vision
setting) and dataset selection (language setting). In both settings,
proxy models are (up to a certain relative scale) effective at approximating the behavior of larger models.

Taking a broader view, many important questions in machine learning reduce to
understanding how changes in training setup (such as training dataset) affect the behvaior of large scale models. Small proxy models can be a powerful tool for practically and effectively answering such questions.

    \section*{Acknowledgment}
    \label{sec:ack}
    Work supported in part by the NSF grant DMS-2134108 and Open Philanthropy.

    \clearpage
    \printbibliography
    \clearpage

    \setcounter{tocdepth}{0}
    \addtocontents{toc}{\protect\setcounter{tocdepth}{3}}

    \appendix

    \clearpage
    \renewcommand{\contentsname}{Appendices}
    \tableofcontents
    \clearpage

    \section{Additional Background}
    \label{sec:app_background}
    In this appendix, we present a more extensive background on datamodels \citep{ilyas2022datamodels} and the corresponding TRAK estimator \citep{park2023trak}. We also present an extensive analysis of the compute requirement for attributing models using TRAK \citep{park2023trak}. We finally present how datamodels \citep{ilyas2022datamodels} could be used to select optimal training sets \citep{engstrom2024dsdm}.

\paragraph{Notation.}
Recall that the training set $S = \{z_1, \ldots, z_n \} \subset \mathcal{Z}$ is a collection of training examples $z_i$ that could be image-label pairs or text samples. Let $L(z; \theta)$ represent the loss of a model with parameters $\theta$ on the example $z_i$. Our models are trained to minimize the empirical risk on the training set, i.e., the parameters $\theta^*(S)$ are computed as follows:
\begin{equation}
    \theta^*(S) := \arg \min_\theta \sum_{z_i \in S} L(z_i; \theta).
    \label{eq:app_opt_params}
\end{equation}

The goal of {\it data attribution} is to trace back a model's prediction to the training data points. Formally, given an example $z$, a training dataset $S$, and a model output function $f(z; \theta)$, a data attribution function $\tau(z; S)$ is function $\tau: \mathcal{Z} \times \mathcal{Z}^n \rightarrow \mathbb{R}^n$ that maps the example $z$ and the training dataset $S$ to a real-valued score vector, called the attribution scores, where the $i^{th}$ entry corresponds to the overall importance of the training example $z_i$ on the model output $f(z; \theta^*(S))$.

\subsection{Datamodels}
\label{sec:app_background_dm}

\subsubsection{Intuition}
\label{sec:app_dm_intro}

As presented in \cref{sec:prelims} of the main paper, datamodels are a tool to approximate how the model output changes when trained on some subset $S'$ of the training set $S$ \citep{ilyas2022datamodels}. Specifically, given a model with parameters $\theta^*(S')$ trained on a training subset $S'$, the goal of datamodels is to approximate how the model output $f(z; \theta^*(S'))$ on example $z$ changes for different subsets $S'$ of the training set $S$. The model output function could represent the loss of the model on the example $z$ or any other metric of interest\footnote{We have presented two different examples of model output functions in the main paper.}.

The model output function $f(z; \theta^*(S'))$ is complex to analyze as it involves training a model on the subset $S'$ and then evaluating the resulting model on the example $z$. Instead, \citet{ilyas2022datamodels} propose approximating this complex function $f(z; \theta^*(S'))$ using a simpler {\it surrogate function} $g(S')$ \citep{designofexperiments} that doesn't involve training a new model. In practice, linear surrogate functions of the form provided a reasonable approximation of the model output \citep{ilyas2022datamodels, saunshi2022understanding}. In particular, for a subset $S'$ of $S$, let $\mathbf{1}_{S'} \in \mathbb{R}^n$ be the indicator vector of $S'$ in $S$, i.e.,
\begin{equation}
    \left( \mathbf{1}(S') \right)_j =
    \begin{cases}
    1  & \text{if $z_j \in S'$} \\
    0 & \text{otherwise}
    \end{cases}
    \label{eq:app_indicator}
\end{equation}
and let $w_{DM} \in \mathbb{R}^n$ be a {\it datamodel vector} (which we explain later how to compute). \citet{ilyas2022datamodels} propose the linear surrogate function
\begin{equation}
    g(S') := \mathbf{1}_{S'}^\top w_{DM}
    \label{eq:app_dm_surrogate}
\end{equation}
to approximate the model output function $f(z; \theta^*(S'))$. The attribution scores are defined as $\tau_{DM}(z; S) = w_{DM}$.

\subsubsection{Computing the Datamodel Vector $w_{DM}$}
\label{sec:app_compute_dm}

A good datamodel vector $w_{DM}$ is one that leads to a surrogate function that approximates well the model output function $f(z; \theta^*(S'))$. When a compute is not an issue, we can search for such a vector using an optimization program that optimizes directly for our goal (good output predictability). This can be achieved as follows:
\begin{enumerate}[label={\bf Step \arabic*.},leftmargin=0.5in]
    \item Sample at random $M$ training subsets $\{S_i: \, S_i \subset S \}_{i=1}^M$ and collect their indicator vectors $\{\mathbf{1}_{S_i}\}_{i=1}^M$.
    \item Train a model on each subset $S_i$ and collect model parameters $\{\theta^*(S_i) \}_{i=1}^M$.
    \item Compute the output of each model for example $z$, i.e., $\{f(z; \theta^*(S_i))\}_{i=1}^M$.
    \item Compute the datamodel vector $w_{DM}$ by regression on the dataset $\left\{\left(\mathbf{1}_{S_i}, \, f(z; \theta^*(S_i)) \right)\right\}_{i=1}^M$.
\end{enumerate}

The regression over the dataset $\left\{\left(\mathbf{1}_{S_i}, \, f(z; \theta^*(S_i)) \right)\right\}_{i=1}^M$ is usually performed using LASSO \citep{ilyas2022datamodels,tibshirani1994regression}, i.e.,
\begin{equation}
    w_{DM} = \arg \min_{w} \frac{1}{M} \sum_{i=1}^{M} \left( \mathbf{1}_{S_i}^\top w_{DM} - f(z; \theta^*(S_i))\right)^2 + \beta \cdot {\lVert w \rVert}_1.
    \label{eq:app_dm_computation}
\end{equation}

This procedure produces a datamodel vector $w_{DM}$ that could be used in the context of the surrogate function $g$ to estimate the output $f(z; \theta^*(S'))$ of a model trained on the subset $S'$, without training the model on $S'$. In the context of data attribution, the datamodels attribution scores correspond to the datamodel vector, i.e., $\tau_{DM}(z, S) = w_{DM}$. We present the full procedure in \cref{alg:app_dm_comp}.

\begin{algorithm}[h]
    \caption{Computing the datamodel vector $w_{DM}$}
    \label{alg:app_dm_comp}
    \begin{algorithmic}[1]
    \Require Target example $z$, dataset $S = \{z_i\}_{i=1}^n$ with $n$ samples, subset ratio $\alpha$, number of models $M$, regularization parameter $\beta$
    \State Sample $M$ random subsets $S_1,S_2,\ldots,S_M \subset S$ of size $\lfloor \alpha \cdot n \rfloor$
    \For{$i \in  1$ to $M$}
        \State Record indicator vector $\mathbf{1}_{S_i}$
        \State Train model on $S_{i}$ and collect parameters $\theta^*(S_i)$
        \State Record the model output function $f(z; \theta^*(S_i))$
    \EndFor

    \State Compute datamodel vector $w_{DM}$ as:
    \begin{equation*}
        w_{DM} = \arg \min_{w} \frac{1}{M} \sum_{i=1}^{M} \left( \mathbf{1}_{S_i}^\top w_{DM} - f(z; \theta^*(S_i))\right)^2 + \beta \cdot {\lVert w \rVert}_1.
    \end{equation*}

    \State
    \Return $w_{DM}$
    \end{algorithmic}
\end{algorithm}

\subsection{Approximating Datamodels with TRAK}
\label{sec:app_trak_approx_dm}

In the following section, we present how TRAK \citep{park2023trak} provides an efficient estimate of datamodels \citep{ilyas2022datamodels}. For a more extensive analysis, please refer to the TRAK paper \citep{park2023trak}.

\subsubsection{Intuition}
\label{sec:app_trak_intro}

Computing the attribution scores using datamodels is an expensive process \citep{ilyas2022datamodels} as it involves training a large number of models $M$ on subsets of the training dataset. This approach is not feasible beyond simple toy settings. To reduce the computational requirement, \citet{park2023trak} propose approximating datamodels by first casting the problem into a logistic regression setup, and then computing the attribution scores efficiently in this new regime. At a high level, casting the original problem into a regression setup can be done by representing the model at hand using a kernel machine \citep{jacot2018neural}. Once the problem is cast into this simple form, prior work has developed a closed-form solution for data attribution in a logistic regressing setup \citep{pregibon1981logistic}. Below, we first present the solution for the logistic regression setup and then present how to cast classification with neural networks into this linear setup.

\subsubsection{Approximating Datamodels in a Logistic Regression Setup}
\label{sec:app_trak_log_reg}

We borrow notation from \citep{park2023trak} and refer the readers to the paper for a more extensive analysis. Consider a logistic regression setup where we have a dataset $S = \{z_1, \ldots, z_n \}$ where each example $z_i = (x_i, b_i, y_i)$ is triple of an input $x_i \in \mathbb{R}^d$, a bias term $b_i \in \mathbb{R}$ and a label $y_i \in \{-1, 1 \}$.

In this setup, we can formulate the logistic regression problem:
\begin{equation}
    \theta^*(S) := \arg \min_\theta \sum_i \log \left[ 1 + \exp (-y_i \cdot (x_i^\top \theta + b_i) ) \right].
    \label{eq:app_log_reg}
\end{equation}
In this simple setup, we define our model output function as the logit function: $f(z; \theta) := x^\top \theta + b$, where $z = (x, b, y)$.

The problem of data attribution in this simple setup is well-studied in literature, and prior work has developed a closed-form solution for it \citep{pregibon1981logistic}. In particular, the contribution of a training example $z_i$ to the model output function $f(z; \theta)$ can be measured using the {\it leave-one-out} influence (LOO) \citep{pregibon1981logistic}, described below:
\begin{equation}
    \tau_{LOO}(z, S) := \frac
    {x^\top (X^\top R X)^{-1} x_i}
    {1 - x_i^\top (X^\top R X)^{-1} x_i \cdot p^*_i \cdot (1-p^*_i)}
    \cdot (1-p^*_i) \approx f(z; \theta^*(S)) - f\left(z, \theta^*(S\backslash \{z_i\}) \right),
    \label{eq:app_loo_logreg}
\end{equation}
where $X \in \mathbb{R}^{n \times d}$ is the matrix of stacked inputs $x_i$, and $p^*_i = \left[ 1 + \exp(-y_i \cdot f(z_i; \theta^*)) \right]^{-1}$ is predicted probability of the correct class, $R \in \mathbb{R}^{n \times n}$ is a diagonal matrix where $R_{ii} = p^*_i \cdot (1-p^*_i)$, and $S \backslash \{z_i\}$ is the training set without example $z_i$. This influence score approximates the effect of removing training example $z_i$ from the training dataset.

In practice, computing the attribution scores in a logistic regression setup using this closed-form solution is efficient and fast. Many interesting problems in ML, however, are highly non-linear. In the next section, we show how we can cast a non-linear problem using neural networks into linear regression problems.

\subsubsection{Casting Non-Linear Problems into Logistic Regression}
\label{sec:app_trak_binary}

In this section, we first start by considering a non-linear binary regression setup. We then present how to generalize the approach to multi-class classification and language modeling.

Given a non-linear binary regression setup, we can express the parameters of the model trained on the dataset as:
\begin{equation}
    \theta^*(S) := \arg \min_\theta \sum_i \log \left[ 1 + \exp (-y_i \cdot f(z_i; \theta) ) \right].
    \label{eq:app_nonlin_log_reg}
\end{equation}

The main challenge in this setup is the non-linearity in the model output function $f(z; \theta)$. \citet{park2023trak} propose to solve this problem by casting the problem at hand into a linear problem. Specifically, given a neural network with model output function $f(z;\theta)$, the authors approximate the model output function around the parameters $\theta^*$ of the optimal model using a Taylor's approximation:
\begin{equation}
    \hat{f}(z; \theta) := f(z;\theta^*) + \nabla_\theta f(z; \theta^*)^\top (\theta - \theta^*).
    \label{eq:app_taylor}
\end{equation}
This step corresponds in the literature to replacing the binary classifier with its eNTK approximation \citep{jacobsen2018irevnet,atanasov2021neural, wei2022more}. Given this linearization, we adapt \cref{eq:app_nonlin_log_reg} and write instead:
\begin{align}
    \theta^*(S) & := \arg \min_\theta \sum_i \log \left[ 1 + \exp (-y_i \cdot f(z_i; \theta) ) \right] \\
    & := \arg \min_\theta \sum_i \log \left[ 1 + \exp \left(-y_i \cdot \left( f(z_i; \theta^*) + \nabla_\theta f(z_i; \theta^*)^\top (\theta - \theta^*) \right) \right) \right] \\
    & := \arg \min_\theta \sum_i \log \left[ 1 + \exp \left(-y_i \cdot \left( \nabla_\theta f(z_i; \theta^*)^\top \theta  + f(z_i; \theta^*) - \nabla_\theta f(z_i; \theta^*)^\top \theta^* \right) \right) \right] \\
    & := \arg \min_\theta \sum_i \log \left[ 1 + \exp \left(-y_i \cdot \left( g_i ^\top \theta + b_i \right) \right) \right],
    \label{eq:app_derive_casting}
\end{align}
where the vector $g_i := \nabla_\theta f(z_i; \theta^*)$ corresponds to the model gradients and we define the bias term $b_i := f(z_i; \theta^*) - \nabla_\theta f(z_i; \theta^*)^\top \theta^*$.

The form we observe in \cref{eq:app_derive_casting} is reminiscent of \cref{eq:app_log_reg}. In fact, given our examples $z_i = (g_i, b_i, y_i)$, we can apply in closed-form the solution from \cref{eq:app_loo_logreg} to compute the attribution scores. However, one big issue in practice is the large dimensionality of the vector $g_i$, which corresponds to the number of model parameters. This value could be in the billions for the largest available models and as such estimating the attribution scores using \cref{eq:app_loo_logreg} is intractable.

\subsubsection{Reducing the Dimensionality and Estimating Datamodels}
\label{sec:app_trak_rand_proj}

Given the intractability of the problem, \citet{park2023trak} propose reducing the dimensionality of the gradient vectors $g_i$ using random projections \citep{johnson1984extensions}. While many techniques exist for reducing the dimensionality of a vector, the authors choose random projections since they preserve some desired properties in the logistic regression problem. We refer the readers to \citep{park2023trak} and \citep{malladi2022kernel} for more details on this choice.

Given a vector $g \in \mathbb{R}^p$ and a random matrix $\mathbf{P} \in \mathbb{R}^{k \times p}$, where $k \ll p$, we define the feature map $\phi: \mathbb{R}^p \rightarrow \mathbb{R}^k$ as $ \phi(g) = \mathbf{P}^\top g$. With this feature map, we project all gradients $g_i$ to obtain feature vectors $\phi_i = \phi(g_i) = \mathbf{P}^\top g_i$, and stack them into the matrix $\Phi := [ \phi_1, \ldots, \phi_n] \in \mathbb{R}^{n \times k}$. Notice how this matrix is much smaller than the original matrix $X = [g_1, \ldots, g_n] \in \mathbb{R}^{n \times p}$.

Using the matrix $\Phi$ of stacked gradients, we can compute the attribution scores as:
\begin{equation}
    \tau_{TRAK}(z, S) = \phi(z)^\top (\Phi^\top \Phi)^{-1} \Phi^\top \mathbf{Q},
    \label{eq:app_trak_single}
\end{equation}
where $\phi(z) = \mathbf{P}^\top \nabla_\theta f(z; \theta^*)$ corresponds to the projected gradient of the target example $z$, and the matrix $\mathbf{Q} := \text{diag}(\{1-p^*_i\}_i)$ is a diagonal matrix with the probabilities of the correct class $p^*_i = \left[ 1 + \exp(-y_i \cdot f(z_i; \theta^*)) \right]^{-1}$. \citet{park2023trak} find that dropping the matrix $R$ and the denominator do not affect the predictiveness of the attribution scores. For more details, we refer the readers to the paper \citep{park2023trak}.

\subsubsection{Improving the Datamodels Estimation using Additional Models}
\label{sec:app_trak_many_models}

One main challenge with the previous approach is the stochastic nature of training models. In particular, changing the random seed and training the same model on the same dataset can lead to widely different results across multiple runs \citep{nguyen2020wide,damour2020underspecification}. To solve this problem, \citet{park2023trak} propose training $M$ models and then averaging across multiple runs as follows:
\begin{equation}
    \tau_{TRAK}(z, S) = \left( \frac{1}{M} \sum_{m=1}^M \phi_m(z)^\top (\Phi_m^\top \Phi_m)^{-1} \Phi_m^\top \right) \cdot \left( \frac{1}{M} \sum_{m=1}^{M} \mathbf{Q}_m \right),
    \label{eq:app_trak_multi_old}
\end{equation}
where the feature map and vectors are different for each of the $M$ runs. Notice that the authors average across the feature maps rather than over attribution scores for numerical stability reasons \citep{park2023trak}.

In this work, we propose a further modification where we drop the term corresponding to the matrix $\mathbf{Q}_m$ from our estimator. Specifically, we compute the attribution scores as:
\begin{align}
    \tau_{TRAK}(z, S) & = \frac{1}{M} \sum_{m=1}^M \phi_m(z)^\top (\Phi_m^\top \Phi_m)^{-1} \Phi_m^\top \\
    & = \frac{1}{M} \sum_{m=1}^M \tau_{TRAK}^{(m)}(z, S).
    \label{eq:app_trak_multi_new}
\end{align}
We notice that dropping the last term does not affect negatively the predictiveness of the attribution scores, and can in many cases in practice improve it. In particular, for many models, the pre-softmax logit can be very large and saturates the softmax when computing probabilities, which in turn leads to multiple 0 entries in the matrix $\mathbf{Q}_m$ and consequently the attribution scores. This behavior reduces drastically the counterfactual predictability, measured using the LDS.

\subsubsection{Generalizing to Multi-Class Classification}
\label{sec:app_trak_multi}

In the previous sections, we presented how to cast general non-linear binary classification problems into a linear regression setup in order to estimate the attribution scores efficiently. In this section, we show how \citet{park2023trak} extended the previous approach to support general multi-class classification setups.

Given a multi-class classification problem over $c$ classes, let $p(z;\theta)$ be the probability assigned by the model to the {\it correct} class. \citet{park2023trak} define the model output function in this setup to be:
\begin{equation}
    f(z;\theta) = \frac{p(z;\theta)}{1-p(z;\theta)}.
    \label{eq:app_multiclass_model_out}
\end{equation}
This model output function essentially measures whether the correct class is more likely than any other class\footnote{This is more tractable than defining $c^2$ classification problems between all pairs of classes.}. One nice property of this model output function is that it allows to write the loss function $L(z; \theta)$ as follows:
\begin{align}
    L(z;\theta) & = - \log(p(z;\theta))  \\
    & = \log \left[ 1 + \exp \left( -f(z;\theta) \right) \right],
    \label{eq:app_multiclass_loss}
\end{align}
which is reminiscent of \cref{eq:app_nonlin_log_reg} (with $y_i=1$). As such, we can make the same approximations made in the binary case setup and apply the same results and derivations to compute the attribution scores. We present the full procedure in \cref{alg:app_trak_comp}.

\begin{algorithm}[h]
    \caption{Approximating the datamodel vector using TRAK for multi-class classification}
    \label{alg:app_trak_comp}
    \begin{algorithmic}[1]
    \Require Target example $z$, dataset $S = \{z_i\}_{i=1}^n$ with $n$ samples, number of models $M$, correct-class likelihood $p(z;\theta)$, projection dimension $k \in \mathbb{N}$
    \State Define model output function: $f(z;\theta) := \frac{p(z;\theta)}{1-p(z;\theta)}$
    \For{$m \in  1$ to $M$}
        \State Train model with parameters $\theta^*_{m}(S)$ on dataset $S$
        \State Sample projection matrix $\mathbf{P}_m \sim \mathcal{N}(0, 1)^{n \times k}$
        \For{$i \in  1$ to $n$}
            \State Compute gradient and project: $\phi_i = \mathbf{P}_m^\top \nabla_\theta f(z_i; \theta^*_m(S))$
        \EndFor
        \State Stack projected gradients: $\Phi_m = [\phi_1, \ldots, \phi_n]^\top$
    \EndFor

    \State Compute the attribution scores using:
    \begin{equation*}
        \tau_{TRAK}(z, S) = \frac{1}{M} \sum_{m=1}^M \phi_m(z)^\top (\Phi_m^\top \Phi_m)^{-1}\Phi_m^\top
    \end{equation*}

    \State
    \Return $\tau_{TRAK}(z, S)$
    \end{algorithmic}
\end{algorithm}

\subsubsection{Adapting the TRAK Estimator to Language Models}
\label{sec:app_trak_nlp}

So far, we have presented how TRAK \citep{park2023trak} could be applied for classification setups. We now present how TRAK could be extended to support language models, as presented in \citep{engstrom2024dsdm}.

Recall that for multi-class classification, \citet{park2023trak} define the model output function to be:
\begin{equation}
    f(z;\theta) = \frac{p(z;\theta)}{1-p(z;\theta)},
\end{equation}
where $p(z;\theta)$ is the probability of the correct class. This setup can be naturally extended to language models trained based on next-token prediction \citep{sutskever2014sequence, vaswani2017attention} where the goal is to iteratively predict out of many tokens the correct token to continue the sentence. Specifically, given a sequence $z = \{z_1, \ldots, z_T \}$ of context length $T$, let $p(z_j \mid z_{< j}; \theta)$ be the probability of predicting the correct token at position $j$ of the sequence, given the previously predicted tokens $z_1, \ldots, z_{j-1}$. This prediction is applied $T-1$ times, with each occurrence being its own classification problem. We can then define the {it language-modeling} model output function as the {\it average} model output function across all classification tasks \citep{engstrom2024dsdm}:
\begin{equation}
    f(z;\theta) = \frac{1}{T} \sum_{j=2}^T \frac{p(z_j \mid z_{<j};\theta)}{1-p(z_j \mid z_{<j};\theta)}.
    \label{eq:app_lm_model_out}
\end{equation}
With this new definition, we can apply the TRAK framework \citep{park2023trak} as outlined in \cref{alg:app_trak_comp}.

\subsection{Estimating Compute Requirement}
\label{sec:app_compute}

In this section, we give an overview of the overall compute requirement. Our analysis focuses mostly on the language setup, where we have observed that compute is a bigger bottleneck. A similar analysis could be done for our vision setup.

\subsubsection{Cost to Train a Single Model}
\label{sec:app_train_cost}

We assume the models being trained are transformers \citep{vaswani2017attention} and leverage the compute approximations presented in \citep{kaplan2020scaling}\footnote{Better approximations exist \citep{hoffmann2022training}, but they do not lead to substantially different approximations.}. Specifically, given a transformer model with $p$ parameters and a dataset composed of $D$ tokens ($n_{train}$ examples\footnote{Given a very large dataset with a total of $n$ examples, compute optimal models can usually be trained using a much smaller number of training examples $n_{train}$\citep{hoffmann2022training}.} with $T$ tokens each), the total cost (measured in FLOPS) for training the transformer on the dataset can be approximated as
\begin{align}
    C^{train} & = C^{forward} + C^{backward} \\
    & = 2pD + 4pD \\
    & = 6p \cdot T \cdot n_{train}.
    \label{eq:app_compute_train}
\end{align}

\subsubsection{Cost to Attribute a Single Model}
\label{sec:app_attrib_cost}

As outlined in the previous sections, the attribution scores (using a single model) on a single target example can be computed using:
\begin{equation}
    \phi(z)^\top (\Phi^\top \Phi)^{-1} \Phi^\top,
    \label{eq:app_trak_bulk}
\end{equation}
where $\phi(z) \in \mathbb{R}^{k}$ is the projected gradient of the target example $z$ and $\Phi \in \mathbb{R}^{n \times k}$ is the stacked matrix of projected inputs, $n$ is the total number of training examples and $k$ is the projection dimension. We assume the cost for multiplying matrices $A \in \mathbb{R}^{a \times b}$ and $B \in \mathbb{R}^{b \times c}$ to be $a \cdot c \cdot (2b-1)$ FLOPS.

We can break down our costs as follows:
\begin{enumerate}
    \item The cost to compute the gradients for the training set is $6pD = 6p \cdot T \cdot n$.
    \item The cost to compute the gradients for the target example is $6p$. When dealing with a target dataset with $n_{test}$ examples, this cost is $6p\cdot T \cdot n_{test}$.
    \item The cost to randomly project the gradients of the training examples is $n\cdot k \cdot (2p-1)$.
    \item The cost to randomly project the gradients of the test examples is $n_{test}\cdot k \cdot (2p-1)$.
    \item The product $\Phi^\top \Phi$ requires $k^2 \cdot (2n - 1)$ FLOPS.
    \item The inverse operation $(\Phi^\top \Phi)^{-1}$ costs around $k^3$ FLOPS.
    \item The product $(\Phi^\top \Phi)^{-1} \Phi^\top$ costs $n \cdot k \cdot (2k - 1)$ FLOPS.
    \item The final product $\phi(z)^\top (\Phi^\top \Phi)^{-1} \Phi^\top$ costs $n\cdot (2k -1)$ FLOPS for a single target example $z$, and $n_{test} \cdot n \cdot (2k-1)$ for attributing over $n_{test}$ target examples.
\end{enumerate}

The total attribution cost is then the sum of the above terms:

\begin{align}
    C^{attrib} & = (6pT + 4k^2 + 2k\cdot p - 2k + 2k \cdot n_{test} - n_{test}) \cdot n \\
    & \qquad + 6p\cdot T \cdot n_{test} + k\cdot(2p-1)\cdot n_{test} - k^2 + k^3 \\
    & \approx (6p \cdot T + 4 k^2 + 2k\cdot p) \cdot n + 2p\cdot (3T + k) \cdot n_{test} + k^3.
    \label{eq:app_compute_attrib}
\end{align}

\subsubsection{Overall Cost}
\label{sec:app_overall_cost}

Using our previous estimates, we can estimate the overall cost as:
\begin{align}
    C^{total} & = C^{train} + C^{attrib} \\
    & = 6p\cdot T \cdot n_{train} + (6p \cdot T + 4 k^2 + 2k\cdot p) \cdot n + 2p\cdot (3T + k) \cdot n_{test} + k^3 \\
    & = \left(6p \cdot T \cdot \left(1 + \frac{n_{train}}{n}\right) + 4 k^2 + 2k \cdot p\right) \cdot n + 2p\cdot (3T + k) \cdot n_{test} + k^3.
    \label{eq:app_compute_overall}
\end{align}

Asymptotically, we find that the ratio of the training cost to the overall cost is:
\begin{align}
    \frac{C^{train}}{C^{total}} & = \frac{6p\cdot n_{train} \cdot T}{\left(6p \cdot T \cdot \left(1 + \frac{n_{train}}{n}\right) + 4 k^2 + 2k \cdot p\right) \cdot n + 2p\cdot (3T + k) \cdot n_{test} + k^3} \\
    & \rightarrow \frac{3\cdot T}{6\cdot T + k} \\
    & \approx 22.22\% \quad \text{(for our setup)},
    \label{eq:app_asymptotic_ratio}
\end{align}
assuming very large compute-optimal models where $n_{train} = n$ \citep{hoffmann2022training}. We present an example of our compute estimates in \cref{tab:app_compute}.

Note that we use $M$ models to improve our attribution scores computed using TRAK \citep{park2023trak}. This increases all our cost estimates by a factor of $M$.

\begin{table}[h]
    \centering
    \caption{Compute requirement for attributing our different MPT models \citep{mosaicml2023llm}.}
    \label{tab:app_compute}
    \vspace{0.1in}
    \begin{tabular}{@{}ccccc@{}}
    \toprule
    {\bf Parameter} & {\bf MPT-125M} & {\bf MPT-350M} & {\bf MPT-760M} & {\bf MPT-8B}\footnote{We do not train an MPT-8B, but instead we include estimates for comparison purposes. MPT-8B corresponds to the largest compute-optimal model that could be trained on our 80 million subset of the MPT dataset.} \\ \midrule
    $p$ ($\times 10^6$ params) & $125$ & $350$ & $760$ & 8000 \\
    $n_{train}$ ($\times 10^6$ examples) & $1.33$ & $3.68$ & $7.47$ & 80 \\
    $n$ ($\times 10^6$ examples) & $80$ & $80$ & $80$ & 80\\
    $n_{test}$ (examples) & 1,000 & 1,000 & 1,000 & 1,000 \\
    $T$ (tokens) & 2,048 & 2,048 & 2,048 & 2,048 \\
    $k$ (proj dim) & 15,360 & 15,360 & 15,360 & 15,360 \\ \midrule
    $C^{train}$ (FLOPS) & $2.04 \times 10^{18}$  & $1.58 \times 10^{19}$ & $6.97 \times 10^{19}$ & $7.86 \times 10^{21}$ \\
    $C^{attrib}$ (FLOPS) & $4.30 \times 10^{20}$ & $1.20 \times 10^{21}$ & $2.62 \times 10^{21}$ & $2.75 \times 10^{22}$ \\
    $C^{overall}$ (FLOPS) & $4.32 \times 10^{20}$ & $1.22 \times 10^{21}$ & $2.68 \times 10^{21}$ &  $3.53 \times 10^{22}$\\
    $\frac{C^{train}}{C^{overall}}$ (\%) & 0.47 & 1.29 & 2.60 & 22.22 \\ \midrule
    $\frac{p}{p\, \text{(MPT-125M)}}$ & 1.00 & 2.80 & 6.08 & 64.00 \\
    $\frac{C^{overall}}{C^{overall} \, \text{(MPT-125M)}}$ & 1.00 & 2.82 & 6.21 & 81.88 \\ \bottomrule
    \end{tabular}
\end{table}

\subsubsection{Practical Considerations}
\label{sec:app_practical_considerations}

In the previous section, we focused solely on the asymptotic behavior. Even in that regime, the boost from using smaller models is already super-linear. In real life, other practical considerations would emerge. For example, models of different sizes might require different amounts of GPU memory, which in turn affects the number of parallel operations within the TRAK framework \citep{park2023trak}. Other considerations include the network bandwidth, especially since we are dealing with massive datasets of several terabytes. All these factors affect our compute estimates and the speedup. The results in \cref{tab:app_compute} merely reflect a lower bound on the speedups in realistic setups.

\clearpage
\subsection{Dataset Selection with Datamodels (\textsc{DsDm})}
\label{sec:app_dsdm}

In this section, we present additional background on the downstream application of data attribution that we consider: dataset selection \citep{brown2020language,xie2023data,engstrom2024dsdm}. We focus on the setup adopted in \citep{engstrom2024dsdm}. For more details, we refer the reader to the paper \citep{engstrom2024dsdm}.

\subsubsection{Problem Setup}
\label{sec:app_dsdm_prob_setup}

Dataset selection refers to the task of selecting from a large pool of data a training set that leads to the ``best'' performance on a given target task. \citet{engstrom2024dsdm} cast the dataset selection task into an optimization problem where the objective function is the model loss on a target task and the decision variable is the dataset selected from a large pool of data.

More precisely, given a large pool of data $\mathcal{Z}$, a target distribution $\mathcal{D}_{\text{targ}}$ (e.g., a language modeling task) and a target dataset size $n$, we can formulate the dataset selection task as:
\begin{align}
    S^* & := \arg \min_{\substack{S \subset \mathcal{Z} \\ \mid S \mid = n}} \mathcal{L}_{\mathcal{D}_{\text{targ}}}(S) \\
    & := \arg \min_{\substack{S \subset \mathcal{Z} \\ \mid S \mid = n}} \mathbb{E}_{z \sim \mathcal{D}_{\text{targ}}} \left[ L(z; \theta^*(S)) \right]
    \label{eq:app_dsdm}
\end{align}
where $\theta^*(S)$ are the parameters of the model trained on $S$, $L(z; \theta^*(S))$ is the loss achieved by the model on target example $z \sim \mathcal{D}_{\text{targ}}$ and $\mathcal{L}_{\mathcal{D}_{\text{targ}}}(S)$ is the expected loss of the models trained on $S$ on samples from the target distribution $\mathcal{D}_{\text{targ}}$.

\subsubsection{Approximating Solution with Datamodels}
\label{sec:app_dsdm_dm}

The optimization problem in \cref{eq:app_dsdm} is generally hard to solve as it involves a combinatorial search over $\footnotesize\begin{pmatrix} \mid\mathcal{Z}\mid \\ n \end{pmatrix}$ possible solutions. Furthermore, evaluating each candidate solution $S$ requires training a new model on the chosen training set $S$ then measuring the model's loss on the target task.

To circumvent this problem, \citet{engstrom2024dsdm} propose using datamodels \citep{ilyas2022datamodels} to approximate the loss of the model trained on the candidate solution $S$. An additional advantage of this approach is the linear relationship between the indicator vector of the set $S$ and the target loss (see \cref{eq:app_dm_surrogate}), which makes the optimization problem easier.

Recall that for a given example $z$, datamodels approximate the complex model output function $f(z; \theta^*(S))$ using a linear surrogate function $g(S) = \mathbf{1}_S ^\top w_z$, where $w_z \in \mathbb{R}^{\mid \mathcal{Z} \mid}$ is the datamodel vector corresponding to target example $z$\footnote{We refer to the datamodel vector $w_{DM}$ as $w$ for ease of notation.}. Using the linear surrogate function, we can approximate for a candidate set $S$ the model's expected loss as:
\begin{align}
    \mathbb{E}_{z \sim \mathcal{D}_{\text{targ}}} \left[ L(z; \theta^*(S)) \right] & \approx \mathbb{E}_{z \sim \mathcal{D}_{\text{targ}}} \left[ \mathbf{1}_S^\top w_z \right] \\
    & = \mathbf{1}_S^\top \mathbb{E}_{z \sim \mathcal{D}_{\text{targ}}} \left[ w_z \right] \\
    & \approx \mathbf{1}_S^\top \left( \frac{1}{m} \sum_{i=1}^m w_{z_i} \right)
    \label{eq:app_dm_approx_loss}
\end{align}
where we assume we have access to $m$ samples from the target distribution $\mathcal{D}_{\text{targ}}$. With this approximation, we rewrite the optimization program from \cref{eq:app_dsdm} as:
\begin{align}
    S^* & := \arg \min_{\substack{S \subset \mathcal{Z} \\ \mid S \mid = n}} \mathbb{E}_{z \sim \mathcal{D}_{\text{targ}}} \left[ L(z; \theta^*(S)) \right] \\
    & \approx \arg \min_{\substack{S \subset \mathcal{Z} \\ \mid S \mid = n}} \mathbf{1}_S^\top \left( \frac{1}{m} \sum_{i=1}^m w_{z_i} \right) \\
    & = \arg \text{bot-}n \left( \frac{1}{m} \sum_{i=1}^m w_{z_i} \right)
    \label{eq:app_dsdm_approx}
\end{align}
which corresponds to choosing the indices corresponding to the bottom $n$ values of the vector $\left( \frac{1}{m} \sum_{i=1}^m w_{z_i} \right)$.

With this new formulation, the task of dataset selection reduces to estimating the datamodels vectors for a given downstream task and then finding the indices corresponding to the bottom $n$ values of the average datamodels vector.

In practice, computing datamodels \citep{ilyas2022datamodels} is expensive, so \citet{engstrom2024dsdm} approximate them using the TRAK framework \citep{park2023trak}. We present an overview of the procedure in \cref{alg:app_dsdm_comp}.

\begin{algorithm}[h]
    \caption{Dataset selection using datamodels (\textsc{DsDm})}
    \label{alg:app_dsdm_comp}
    \begin{algorithmic}[1]
    \Require Large pool of data $\mathcal{Z}$, selected dataset size $n$, $m$ target examples $\{z_1, \ldots, z_m\}$ from distribution $\mathcal{D}_{\text{targ}}$
    \State Estimate datamodels vectors $\{w_{z_i} \}_{i=1}^m$ from $\mathcal{Z}$ using TRAK
    \State Compute average datamodel vector $w_{\text{targ}} = \left( \frac{1}{m} \sum_{i=1}^m w_{z_i} \right)$
    \State Collect indices $\mathcal{I} = \arg \text{bot-}n \left( w_{\text{targ}} \right)$

    \State
    \Return optimal set $S^*$ of training examples from pool $\mathcal{Z}$ at indices $\mathcal{I}$
    \end{algorithmic}
\end{algorithm}

    \clearpage
    \section{Similarity Between Small and Large Models}
    \label{sec:app_sim}
    In the main paper, we demonstrated that when models of different sizes are trained on the same data distribution, their losses are surprisingly linear (see \cref{fig:small_large_losses_nlp}). In this section, we present additional details on the experimental setup of our result from \cref{fig:small_large_losses_nlp}, and then present more results in the vision setting.

\subsection{Language Setting}

\subsubsection{Experimental Setup}
\label{sec:app_sim_exp_setup}

\paragraph{Models.}
In this setting, we consider two models based on the MPT architecture \citep{mosaicml2023llm}: a small model with 80M parameters and a larger one with 760M parameters. The small model is trained on 1.67B tokens while the large model is trained on 15.3B tokens\footnote{The number of tokens was chosen to optimize for the compute level, as described in \citep{hoffmann2022training}.}. This makes the small model require 85x less compute than the larger model. Both models have a context length of 1,024. More architectural details in \cref{tab:app_small_large_nlp_hparams} below.

\begin{table}[h]
    \centering

    \caption{The architecture of our small and large MPT models \citep{mosaicml2023llm} used for \cref{fig:small_large_losses_nlp}.}
    \label{tab:app_small_large_nlp_hparams}
    \vspace{0.1in}

    \begin{tabular}{@{}lccccc@{}}
    \toprule
                      & \textbf{Model Dim} & \textbf{Heads} & \textbf{Layers} & \textbf{Parameters} & \textbf{Train Tokens (B)} \\ \midrule
    \textbf{MPT-80M} ({\it small}) & 640                & 10             & 10              & 82,127,360 & 1.67         \\
    \textbf{MPT-760M} ({\it large}) & 1,536              & 12             & 24              & 760,470,528 & 15.3        \\ \bottomrule
    \end{tabular}
\end{table}

\paragraph{Data distributions.}
We train several copies of the small and large models, each on a different data distribution. Some of our distributions are {\it natural} while the rest are induced by algorithms.

\begin{itemize}
    \item {\it Natural distributions}:
    \begin{itemize}
        \item {\bf MPT dataset} \citep{mosaicml2023introducing}: The MPT dataset is a collection of examples from several online sources such as CommonCrawl, RedPajama, etc.\footnote{An extensive list of sources can be found at \citep{mosaicml2023introducing}.} We train our models on random subsets from the MPT dataset.
        \item {\bf RedPajama-ArXiV} \citep{together2023redpajama}: The data consists of ArXiV articles and is extracted from the MPT subset.
        \item {\bf RedPajama-Books} \citep{together2023redpajama}: The data consists of subsets of books and is extracted from the MPT subset.
        \item {\bf RedPajama-Wiki} \citep{together2023redpajama}: The data consists of Wikipedia articles and is extracted from the MPT subset.
        \item {\bf Semantic Scholar} \citep{lo2020s2orc}: The data consists of Semantic Scholar articles and is extracted from the MPT subset.
        \item {\bf Stack-Markdown} \citep{kocetkov2022stack}: The data consists of Markdown code from the Stack dataset and is extracted from the MPT subset.
    \end{itemize}

    \item {\it Algorithm-induced distributions}\footnote{The data can be found at \url{https://github.com/MadryLab/DsDm}.}:
    \begin{itemize}
        \item \textsc{\bf DsDm} \citep{engstrom2024dsdm}: \textsc{DsDm} is a method for selecting pretraining examples that improve the downstream performance. We reuse the outcomes of this method when applied to the C4 dataset \citep{raffel2020exploring} as presented in \citep{engstrom2024dsdm}.
        \item \textsc{\bf Bot-DsDm} \citep{engstrom2024dsdm}: This method is simply the reverse of {\sc DsDm}. Specifically, we choose the pretraining examples that hurt performance the most. While this distribution is not particularly useful practically, it is helpful insofar as it reflects how language models behave at the other end of the spectrum.
        \item \textsc{\bf DSIR} \citep{xie2023data}: DSIR is a method to choose pretraining examples that improve performance through importance resampling. We reuse the outcomes of this method when applied to the C4 dataset \citep{raffel2020exploring} as presented in \citep{engstrom2024dsdm}.
        \item \textsc{\bf Classifier} \citep{brown2020language}: Classifier is a method to choose pretraining examples that improve performance by using a classifier that predicts whether the pretraining examples are similar to the downstream examples or not. We reuse the outcomes of this method when applied to the C4 dataset \citep{raffel2020exploring} as presented in \citep{engstrom2024dsdm}.
    \end{itemize}
\end{itemize}

\paragraph{Downstream datasets.}
After training our models on each of the data distributions highlighted above, we measure their losses on several datasets. The goal is to reflect the linearity over multiple data distributions.
\begin{itemize}
    \item {\bf C4} \citep{raffel2020exploring}: This dataset consists of web-extracted text from Common Crawl during April 2019.
    \item {\bf The Pile} \citep{gao2020pile}: This dataset consists of text extracted from multiple sources, including Common Crawl, Books, etc. More details can be found in the paper.
    \item {\bf SQuAD} \citep{rajpurkar2016squad}: Stanford Question Answering Dataset \citep{rajpurkar2016squad} is a reading comprehension dataset composed of excerpts from Wikipedia articles. The task in this dataset is answering questions given some context.
    \item {\bf LAMBADA} \citep{paperno2016lambada}: LAnguage Modeling Broadened to Account for Discourse Aspects \citep{paperno2016lambada} is a dataset that measures broad context understanding through the means of word prediction. \citet{paperno2016lambada} collected text narratives where human annotators are able to predict the last word in a sentence when they have seen the whole passage but not when they only see the last sentence before the text completion. We use the version of the dataset cleaned by EleutherAI\footnote{The dataset can be found on: \url{https://huggingface.co/datasets/EleutherAI/lambada_openai/viewer/en}.}.
    \item {\bf HellaSwag} \citep{zellers2019hellaswag}: HellaSwag is multiple-choice dataset extracted from the SWAG dataset \citep{zellers2018swag}. The dataset is extracted using Adversarial Filtering (AF) and is challenging to language models while being almost trivial for humans.
\end{itemize}

For the downstream datasets SQuAD, LAMBADA and HellaSwag, we measure the models' losses only over their predictions, while for the pretraining datasets C4 and The Pile, we measure their losses over the whole sequence.

\subsubsection{Correlation at the Example Level}

In \cref{fig:small_large_losses_nlp}, we show that the losses achieved by the small and large models on a target distribution are linear. We extend this result and show that for some sequences, in the downstream tasks, the losses achieved at the example level are also linear (see \cref{fig:small_large_losses_seq_57m}).

We then plot the coefficient of determination ($R^2$) between the losses achieved by the small and large models on target examples in each downstream task (see \cref{fig:small_large_r2_dist}). We can see that a significant proportion of the target examples have a positive $R^2$.

\subsubsection{Correlation for Larger Compute Gap}
We now investigate how our results change when we increase the compute gap. To that end, we consider a smaller model consisting of 37 million parameters and trained on 840 million tokens. We consider the same large model of 760 million parameters. The difference in compute in this case is 370x. For more architectural details, check \cref{tab:app_small_large_nlp_hparams_40m}.

\begin{table}[h]
    \centering

    \caption{The architecture of our small and large MPT models \citep{mosaicml2023llm} used for \cref{fig:small_large_losses_nlp}.}
    \label{tab:app_small_large_nlp_hparams_40m}
    \vspace{0.1in}

    \begin{tabular}{@{}lccccc@{}}
    \toprule
                      & \textbf{Model Dim} & \textbf{Heads} & \textbf{Layers} & \textbf{Parameters} & \textbf{Train Tokens (B)} \\ \midrule
    \textbf{MPT-37M} ({\it small}) & 384                & 6             & 10              & 37,479,936 & 0.84         \\
    \textbf{MPT-760M} ({\it large}) & 1,536              & 12             & 24              & 760,470,528 & 15.9        \\ \bottomrule
    \end{tabular}
\end{table}

\begin{figure}[h]

    \centering
    \includegraphics[width=\linewidth]{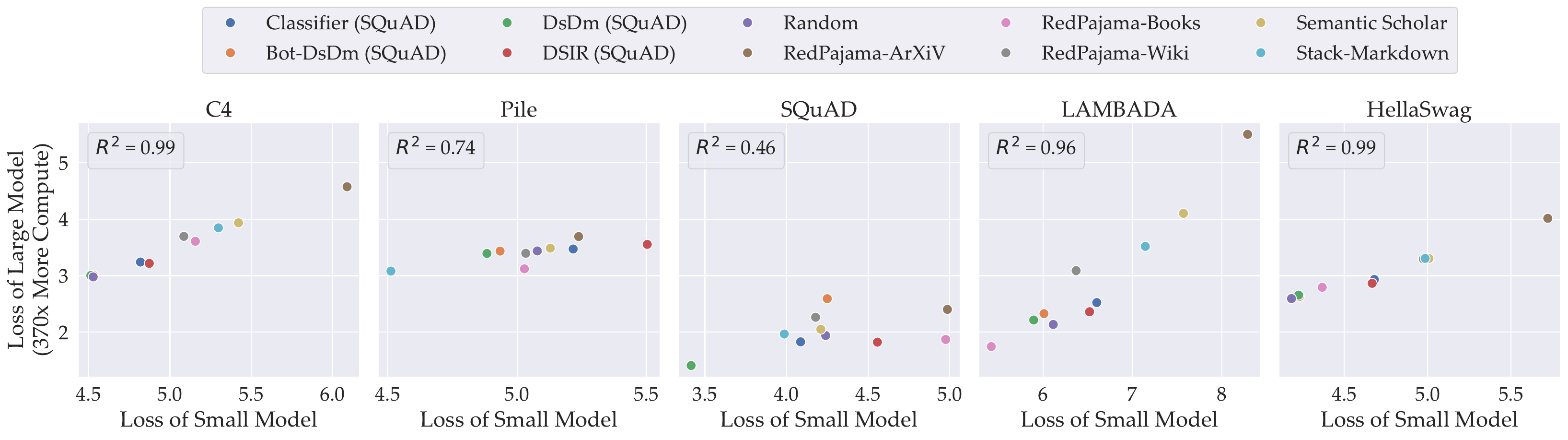}

    \caption{
        Small models are reliable proxies of large models. In all plots, the $x$-axis represents the loss achieved by a small MPT model of 37 million parameters trained on 0.84B tokens and the $y$-axis represents the loss achieved by a larger MPT model of 760 million parameters trained on 15.9B tokens. Each plot corresponds to a different target distribution, and within each plot, each point corresponds to a different training distribution.
    }
    \label{fig:small_large_losses_nlp_40m}

\end{figure}

When the compute gap is larger, we still observe a very strong correlation between small and large models, albeit slightly weaker on some downstream tasks, e.g., SQuAD \citep{rajpurkar2016squad} (see \cref{fig:small_large_losses_nlp_40m}).

\begin{figure}[h]

    \centering
    \includegraphics[width=\linewidth]{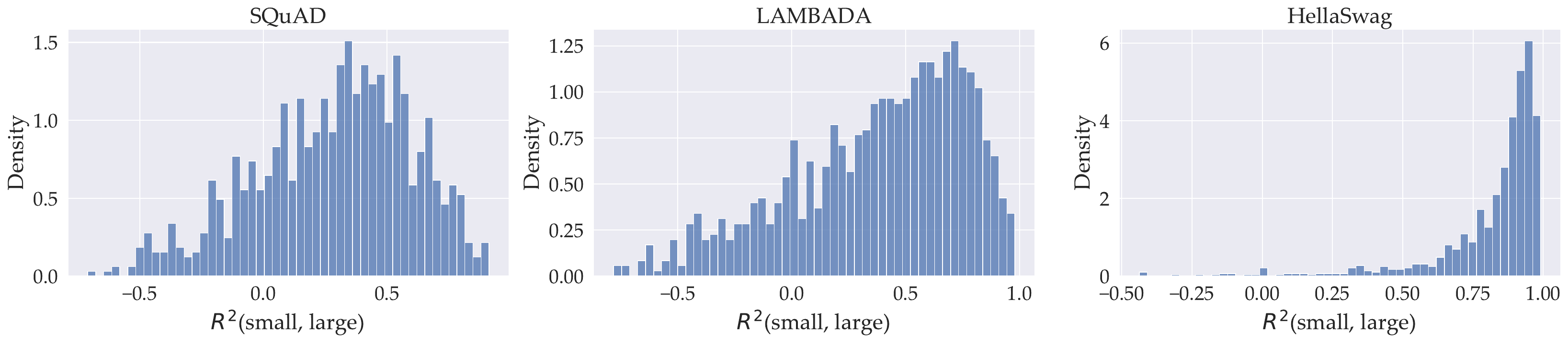}

    \caption{
        Small models are reliable proxies of large models. We plot the coefficient of determination ($R^2$) between the losses of the small and large models for all examples.
    }
    \label{fig:small_large_r2_dist_40m}

\end{figure}

We next investigate how this correlation changes at the example level. Similar to the earlier setting, we observe a strong (still weaker) correlation (see \cref{fig:small_large_r2_dist_40m} and \cref{fig:small_large_losses_seq_57m}). These results indicate that small models can still be reliable proxies of large models, even when the difference in compute is different by orders of magnitude.

\subsubsection{How Correlation Changes with Compute}
\label{sec:app_r2_vs_compute}

We observe a large correlation between the losses achieved by our small and large models over multiple tasks. To test the extent of this correlation, we train several other models of different sizes (125M, 220M and 350M) with different compute budgets, measure the coefficient of determination between their losses and the loss of the large model (760M) and then plot how this correlation (averaged over multiple tasks) changes as a function of the training compute budget. We see in \cref{fig:r2_vs_compute} that the correlation increases as the training compute increases.

\subsection{Vision Setting}

We show in this section that our results still hold across the vision setting.

\subsubsection{Experimental Setup}

\paragraph{Models.}
We consider variants of the ResNet-18 architecture \citep{he2015residual} where we vary the width by a multiplicative factor. Specifically, our small model is ResNet-18 where the width is multiplied by $\frac{1}{4}$ and our large model is a ResNet-18 where the width is multiplied by 2. We provide more information on the architecture in \cref{tab:app_k_vs_nparams}.

\paragraph{Data distributions.}
The dataset we consider for the vision setting is the CIFAR-10 dataset \citep{krizhevsky2009learning}. We track in this setting the margin (instead of the loss) of the small and large model on a specific example (rather than the average margin over the dataset). We choose 4 test examples at random, and for each example, we train and average 8 models on each of the following distributions:

\begin{itemize}
    \item {\bf Random}: We remove at random up to 10\% of the training examples.
    \item {\bf Top Infl}: We estimate using TRAK the influence of every training example on the selected test example \citep{park2023trak,ilyas2022datamodels}, then we create several training datasets where we remove at random up to 10\% of the training examples with the top datamodels score.
    \item {\bf Bot Infl}: We estimate using TRAK the influence of every training example on the selected test example \citep{park2023trak,ilyas2022datamodels}, then we create several training datasets where we remove at random up to 10\% of the training examples with the bottom datamodels score.
    \item {\bf Most Sim}: We compute the similarity (in feature space) of each training example and the selected test example, then we create several training datasets where we remove at random up to 10\% of the training examples the most similar to the selected test examples.
    \item {\bf Least Sim}: We compute the similarity (in feature space) of each training example and the selected test example, then we create several training datasets where we remove at random up to 10\% of the training examples the least similar to the selected test examples.
    \item {\bf Same Class}: For each test example, we remove at random up to \{25\% -- 50\% -- 75\%\} of the training examples from the same class.
\end{itemize}

\subsubsection{Results}

\begin{figure}[t]

    \centering
    \includegraphics[width=\linewidth]{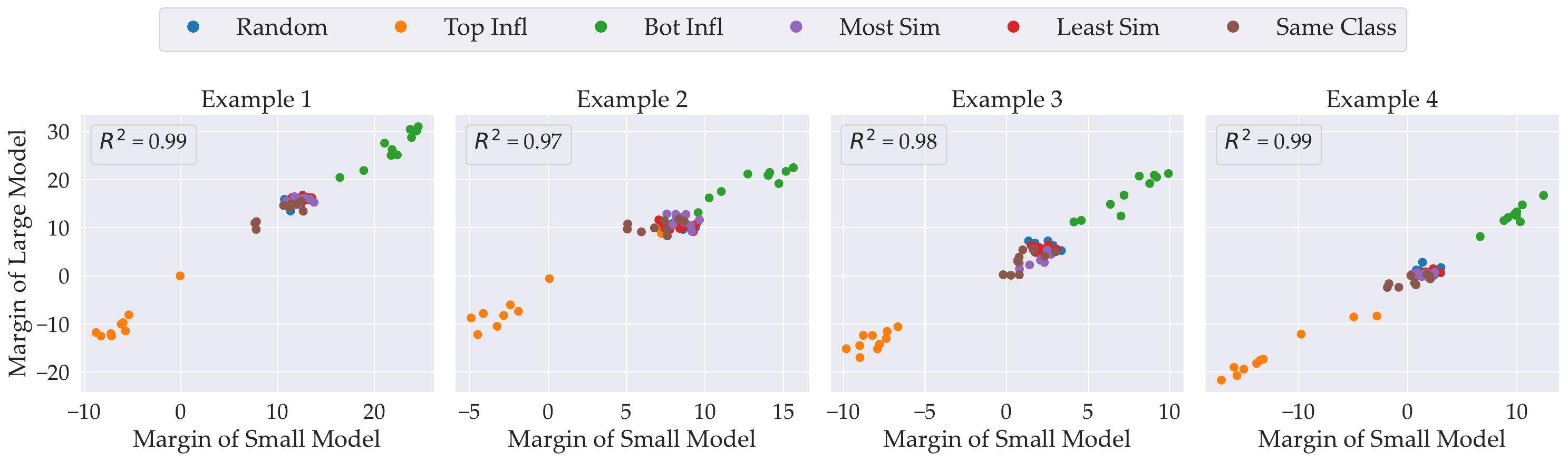}

    \caption{
        Small models are reliable proxies of large models. In all plots, the $x$-axis represents the margin of the small ResNet-18-$\frac{1}{4}$ model and the $y$-axis represents the margin of the larger ResNet-18-2 model. Each plot corresponds to a different test example, and within each plot, each point corresponds to a different training distribution.
    }
    \label{fig:small_large_losses_vision}

\end{figure}

For each test example and each training distribution, we train 8 of the small and large models and record their margins on the selected test example. We see in \cref{fig:small_large_losses_vision} that the margins of the small and large models are linear over the different training distributions.

    \clearpage
    \section{Experimental Setup}
    \label{sec:app_setup}
    In this appendix, we present additional details about our experimental setup.

\subsection{Vision Setup}
\label{sec:app_vision_setup}

\subsubsection{Datasets}
In the vision setup, we consider two small datasets: CIFAR-10 and CIFAR-100 \citep{krizhevsky2009learning} and a larger dataset: ImageNet \citep{krizhevsky2012imagenet}. Both CIFAR datasets are composed of 50,000 training examples and 10,000 test examples belonging to 10 and 100 classes respectively, while ImageNet \citep{krizhevsky2012imagenet} contains 1.2M training examples and 50,000 test examples belonging to 1,000 classes.

\subsubsection{Models}
As presented in the main paper, we consider in the vision setup ResNet-18 models \citep{he2015residual} where we multiply the width of each layer by a factor $k$ and refer to the resulting model as RN-$k$. In the context of ResNets \citep{he2015residual}, the width of a layer refers to the number of output channels in this layer. When the factor $k$ is larger than 1, the model at hand corresponds to a WideResNet-18 \citep{zagoruyko2016wide}. We present in \cref{tab:app_k_vs_nparams} how the model size changes as we increase the width of the network.

\begin{table}[h]

    \centering

    \caption{Number of parameters in each of our models RN-$k$. The difference observed between the CIFAR \citep{krizhevsky2009learning} and ImageNet \citep{russakovsky2015imagenet} datasets corresponds to the difference in the input image size (32 and 224 respectively).}
    \label{tab:app_k_vs_nparams}
    \vspace{0.1in}

    \resizebox{\textwidth}{!}{%
        \begin{tabular}{@{}lcccccccc@{}}
        \toprule
                        & \textbf{1/16} & \textbf{1/8} & \textbf{1/4} & \textbf{1/2} & \textbf{1} & \textbf{2} & \textbf{4}  & \textbf{8}  \\ \midrule
        \textbf{CIFAR}    & 44,622        & 176,402      & 701,466      & 2,797,610    & 11,173,962 & 44,662,922 & 178,585,866 & 714,421,850 \\
        \textbf{ImageNet} & -             & 241,712      & 831,096      & 3,055,880    & 11,689,512 & 45,693,032 & 180,645,096 & -           \\ \bottomrule
        \end{tabular}%
    }
\end{table}

\subsubsection{Training details}
\begin{figure}[h]

    \centering
    \includegraphics[width=\linewidth]{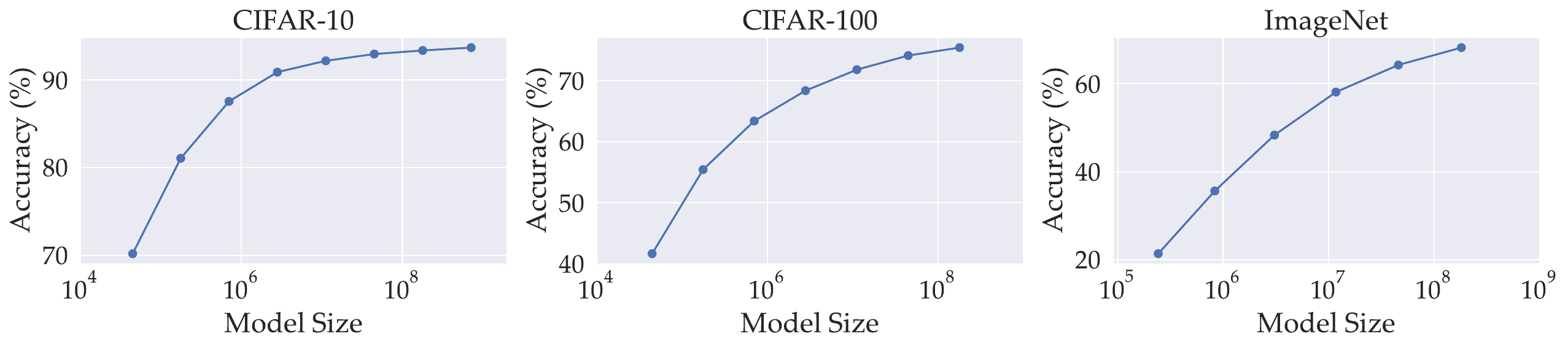}

    \caption{Performance of our models on CIFAR-10, CIFAR-100 \citep{krizhevsky2009learning} and ImageNet \citep{krizhevsky2012imagenet} for different widths.}
    \label{fig:app_vis_model_vs_size}

\end{figure}

We train all our models using the same set of hyperparameters, presented in \cref{tab:app_rn_hparams}. To ensure that our hyperparameters are compatible with all our models of different sizes, we leverage the $\mu P$ framework \citep{yang2022tp5} in our implementation\footnote{We integrate the $\mu P$ GitHub library in our code: \url{https://github.com/microsoft/mup}.}. We refer the readers to \citep{yang2022tp5} for more details on how the $\mu P$ framework works. We show how the accuracy of our model changes as we increase the width in \cref{fig:app_vis_model_vs_size}

\begin{table}[h]

    \centering

    \caption{Hyperparameters used to train our RN-$k$ models. We leverage the $\mu P$ framework \citep{yang2022tp5} in order to use the same hyperparameters for all our models of different sizes.}
    \label{tab:app_rn_hparams}
    \vspace{0.1in}

    \begin{tabular}{@{}lcc@{}}
        \toprule
        \textbf{Hyperparameter} & {\bf CIFAR} \citep{krizhevsky2009learning} & {\bf ImageNet} \citep{krizhevsky2012imagenet} \\ \midrule
        Optimizer               & SGD            & SGD \\
        LR Scheduler            & OneCycle       & OneCycle \\
        Max LR                  & 0.1            &  0.5 \\
        Initial LR              & 0.001          &  0.005 \\
        LR Decay                & Linear         &  Cosine \\
        Warmup (\%)             & 0.05           &  0.05 \\
        Epochs                  & 30             &  20 \\
        Batch Size              & 512            &  512 \\
        Weight Decay            & 0.0005          &  0.0005 \\ \bottomrule
    \end{tabular}

\end{table}

\subsubsection{TRAK details}
\label{sec:app_vis_trak_details}
In this setup, we train 8 independent models RN-$k$ models for each multiplicative factor $k$. We then pass the model checkpoints $\{20\ldots 30\}$ for CIFAR \citep{krizhevsky2009learning} and checkpoints $\{10\ldots 20\}$ for ImageNet \citep{krizhevsky2012imagenet} to the TRAK code. As presented in \cref{sec:app_background}, one important parameter of TRAK \citep{park2023trak} is the projection dimension that corresponds to the dimension of the subspace onto a model's gradients are mapped. The choice of this parameter presents naturally a trade-off between thq quality of the attribution scores and throughput \citep{park2023trak}: increasing the projection dimension increases simultaneously the quality of the attribution scores and the time to compute them. For our setup, we choose projection dimensions of 2,048, 4,096 and 15,360 on CIFAR-10, CIFAR-100 and ImageNet respectively \citep{krizhevsky2009learning, russakovsky2015imagenet}.

The attribution scores that we compute are matrices of $50,000 \times 10,000$ for CIFAR \citep{krizhevsky2009learning} and $1.2M \times 50,000$ for ImageNet \citep{krizhevsky2012imagenet}.

\subsection{Language Setup}
\label{sec:app_lang_setup}

\subsubsection{Datasets}
\label{sec:app_lang_setup_datasets}

\paragraph{Pretraining dataset.}
In the language setup, we consider a large pretraining dataset composed of 80 million samples (subset of the MPT dataset introduced in \citep{mosaicml2023introducing}). We pre-tokenize this dataset before training using the GPT-NeoX tokenizer \citep{gpt-neox-library} (with a vocabulary size of 50,368 tokens). The resulting pre-tokenized dataset contains 160B tokens.

\paragraph{Downstream datasets.}
We consider two downstream datasets for our application: LAMBADA \citep{paperno2016lambada} and SQuAD \citep{rajpurkar2016squad}:
\begin{itemize}
    \item {\bf LAMBADA}: LAnguage Modeling Broadened to Account for Discourse Aspects \citep{paperno2016lambada} is a dataset that measures broad context understanding through the means of word prediction. \citep{paperno2016lambada} collected text narratives where human annotators are able to predict the last word in a sentence when they have seen the whole passage but not when they only see the last sentence before the text completion. We use the version of the dataset cleaned by EleutherAI\footnote{The dataset can be found on: \url{https://huggingface.co/datasets/EleutherAI/lambada_openai/viewer/en}.}. Similar to \citep{engstrom2024dsdm}, we split the dataset into a holdout set of 2,570 samples and a target set of 2,577 samples.
    \item {\bf SQuAD}: Stanford Question Answering Dataset \citep{rajpurkar2016squad} is a reading comprehension dataset composed of excerpts from Wikipedia articles. The task in this dataset is answering questions given some context. Similar to \citep{engstrom2024dsdm}, we split the dataset into a holdout set of 10,557 samples (corresponding to the SQuAD validation set) and a target set of 23,107 examples (corresponding to 25\% of the SQuAD training set).
\end{itemize}

\begin{figure}[t]
    \centering
    \begin{minipage}{\linewidth}
                \small
                \raggedright
                \begin{enumerate} %
                \item {\bf Context}: Formed in 1946, Sierra Sky Park Airport is a residential airport community born of a unique agreement in transportation law to allow personal aircraft and automobiles to share certain roads. Sierra Sky Park was the first aviation community to be built[citation needed] and there are now numerous such communities across the United States and around the world. Developer William Smilie created the nation's first planned aviation community. Still in operation today, the public use airport provides a unique neighborhood that spawned interest and similar communities nationwide.\\
                {\bf Question}: What is the name of the first aviation community built?\\
                {\bf Answer}: \hl{Sierra Sky Park}
                \item {\bf Context}: The Newcastle Beer Festival, organized by CAMRA, takes place in April. In May, Newcastle and Gateshead host the Evolution Festival, a music festival held on the Newcastle and Gateshead Quaysides over the Spring bank holiday, with performances by acts from the world of Rock, Indie and Dance music. The biennial AV Festival of international electronic art, featuring exhibitions, concerts, conferences and film screenings, is held in March. The North East Art Expo, a festival of art and design from the regions professional artists, is held in late May. EAT! NewcastleGateshead, a festival of food and drink, runs for 2 weeks each year in mid June.\\
                {\bf Question}: What festival takes place in April in Newcastle?\\
                {\bf Answer}: \hl{The Newcastle Beer Festival}
                \end{enumerate}
    \end{minipage}
    \caption{Random SQuAD samples \citep{rajpurkar2016squad}. Context is normal text, and
    the continuation label is \hl{hightlighted}.}
    \label{fig:app_squad_samples}
\end{figure}
\begin{figure}[t!]
    \centering
    \begin{minipage}{\linewidth}
                \small
                \raggedright
                \begin{enumerate} %
                \item {\bf Context}: In 1854 at Ballarat there was an armed rebellion against the government of Victoria by miners protesting against mining taxes (the "Eureka Stockade"). This was crushed by British troops, but the discontents prompted colonial authorities to reform the administration (particularly reducing the hated mining licence fees) and extend the franchise. Within a short time, the Imperial Parliament granted Victoria responsible government with the passage of the Colony of Victoria Act 1855. Some of the leaders of the Eureka rebellion went on to become members of the Victorian Parliament.\\
                {\bf Question}: What did colonial authorities  reduce because of the Ballarat revolt?\\
                {\bf Answer}: \hl{mining licence fees}
                \item {\bf Context}: Within southern California are two major cities, Los Angeles and San Diego, as well as three of the country's largest metropolitan areas. With a population of 3,792,621, Los Angeles is the most populous city in California and the second most populous in the United States. To the south and with a population of 1,307,402 is San Diego, the second most populous city in the state and the eighth most populous in the nation.\\
                {\bf Question}: What is the population of Los Angeles?\\
                {\bf Answer}: \hl{3,792,621}
                \end{enumerate}
    \end{minipage}
    \caption{Random LAMBADA samples \citep{paperno2016lambada}. Context is normal text, and
    the continuation label is \hl{hightlighted}.}
    \label{fig:app_squad_samples}
\end{figure}

\subsubsection{Models}
In this setup, we consider three MPT models presented in \citep{mosaicml2023llm}\footnote{We use the code provided in \url{https://github.com/mosaicml/llm-foundry}.}. Our three models are of sizes 125M, 350M and 760M parameters respectively. We present the architecture of the models in \cref{tab:app_mpt_arch}.

\begin{table}[h]
    \centering

    \caption{The architecture and hyperparameters of our three MPT models \citep{mosaicml2023llm}.}
    \label{tab:app_mpt_arch}
    \vspace{0.1in}

    \resizebox{\linewidth}{!}{%
        \begin{tabular}{@{}lcccccccc@{}}
        \toprule
                        & \textbf{Model Dim} & \textbf{Heads} & \textbf{Layers} & \textbf{Parameters} & \textbf{LR} & \textbf{wd} & \textbf{Batch} & \textbf{Total (tokens)} \\ \midrule
        \textbf{MPT-125M} & 768                & 12             & 12              & 125,311,488 & $6 \times 10^{4}$ & $4 \times 10^{-4}$ & 2M & 2.7B        \\
        \textbf{MPT-350M} & 1,024              & 16             & 24              & 355,985,408 & $6 \times 10^{4}$ & $4 \times 10^{-4}$ & 2M & 7.5B        \\
        \textbf{MPT-760M} & 1,536              & 12             & 24              & 760,470,528 & $6 \times 10^{4}$ & $4 \times 10^{-4}$ & 2M & 15.3B         \\ \bottomrule
        \end{tabular}%
    }
\end{table}

\subsubsection{Training details}
\label{sec:app_lang_training}
\begin{figure}[h]

    \centering
    \includegraphics[width=0.5\linewidth]{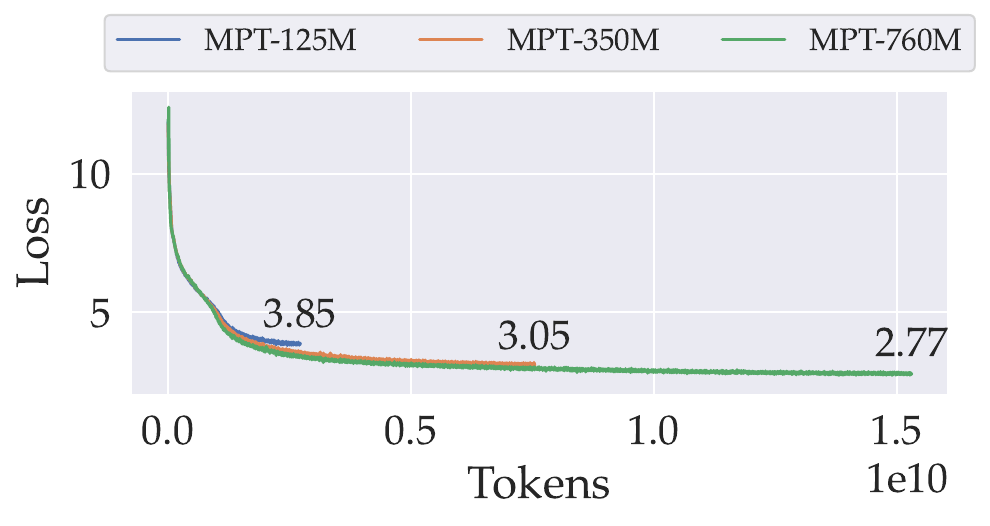}

    \caption{Performance of our three compute-optimal MPT models \citep{mosaicml2023llm,hoffmann2022training}.}
    \label{fig:app_lang_loss_vs_size}

\end{figure}

We train our MPT models using the {\it llm-foundry} repository\footnote{\url{https://github.com/mosaicml/llm-foundry}.} developed by MosaicML on our subset of the MPT dataset \citep{mosaicml2023introducing}. We present some of the hyperparameters used for training our models in \cref{tab:app_mpt_arch}. When training, our models, we pack the tokens from our pre-tokenized dataset into samples of context length 2,048. For the rest of the training hyperparameters, we keep the original values used in the GitHub repository. We show the three training curves of our models in \cref{fig:app_lang_loss_vs_size}.

\subsubsection{TRAK details}
\label{sec:app_lang_trak_details}
In this setup, the computational requirement is much higher. For that reason, we only train three different models of each size on different random subsets of the training dataset (see \cref{tab:app_mpt_arch} for the total number of tokens of each model). We then pass to TRAK these three checkpoints in order to compute the attribution scores of all our training examples. In this setup, we use a projection dimension of 15,360.

In the language setup, TRAK produces for each of our three models two sets of attribution scores: one for LAMBADA \citep{paperno2016lambada} and the other for SQuAD \citep{rajpurkar2016squad}, each computed using the samples from the target set (see \cref{sec:app_lang_setup_datasets}). The attribution scores we compute are vectors containing 80 million entries (one for each training example).

\subsection{Dataset Selection}
\label{sec:app_dsdm_setup}

For this downstream application, we compute the attribution scores as outlined in \cref{sec:app_lang_trak_details} (based on the target set of each dataset) and then we train our large models (MPT-760M \citep{mosaicml2023llm}) on the selected dataset, using the recipe described in \cref{sec:app_lang_training}. We test the performance of our models on the holdout sets of each dataset.

    \clearpage
    \section{Additional Results}
    \label{sec:app_results}
    \subsection{Qualitative Similarity}
\label{app:results_qual_sim}

\subsubsection{Vision Setup}
\label{app:results_qual_sim_vis}

\begin{figure}[h]

    \centering
    \includegraphics[width=0.9\linewidth]{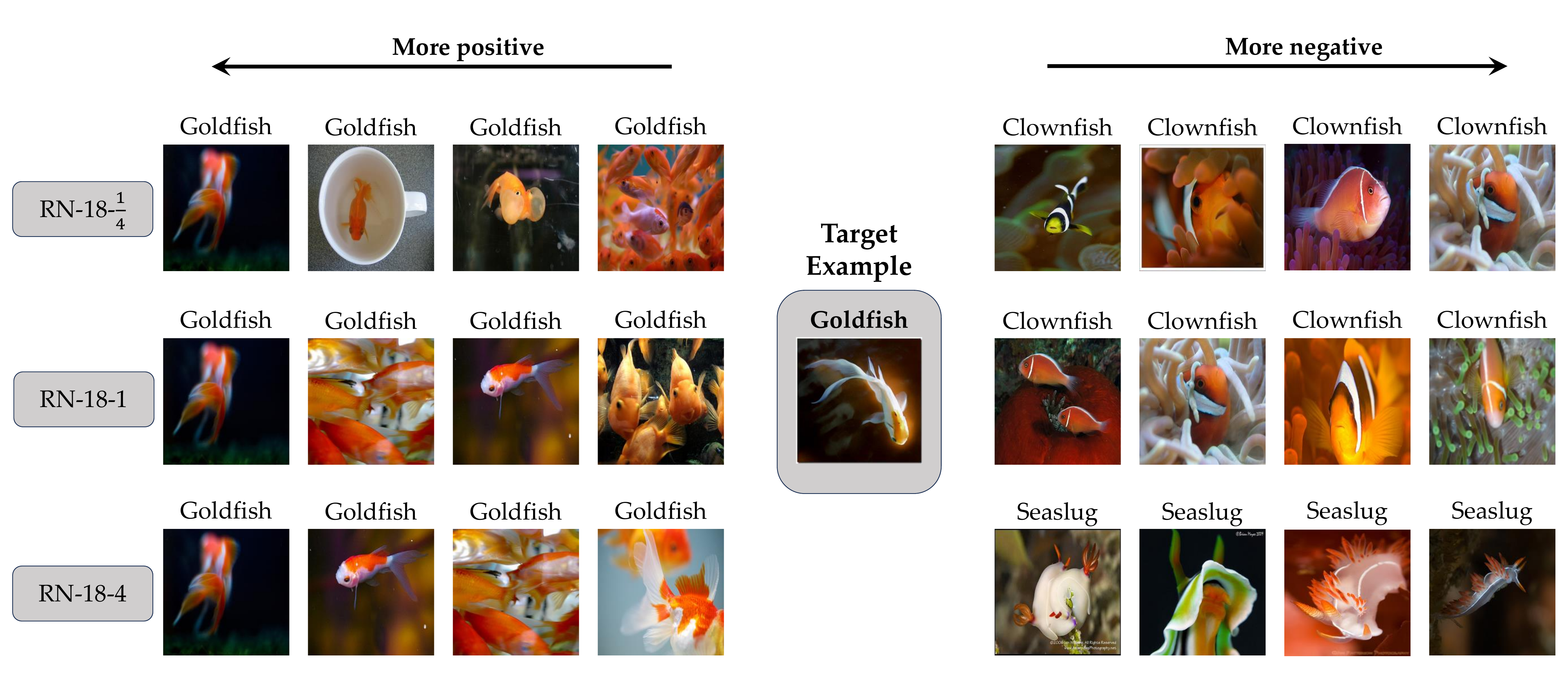}

    \caption{Most helpful and detrimental examples for the outputs of models of different sizes are similar. We observe a large overlap between the examples that are most helpful (and most detrimental) for the models predictions on the target example.}

\end{figure}
\begin{figure}[h]

    \centering
    \includegraphics[width=0.9\linewidth]{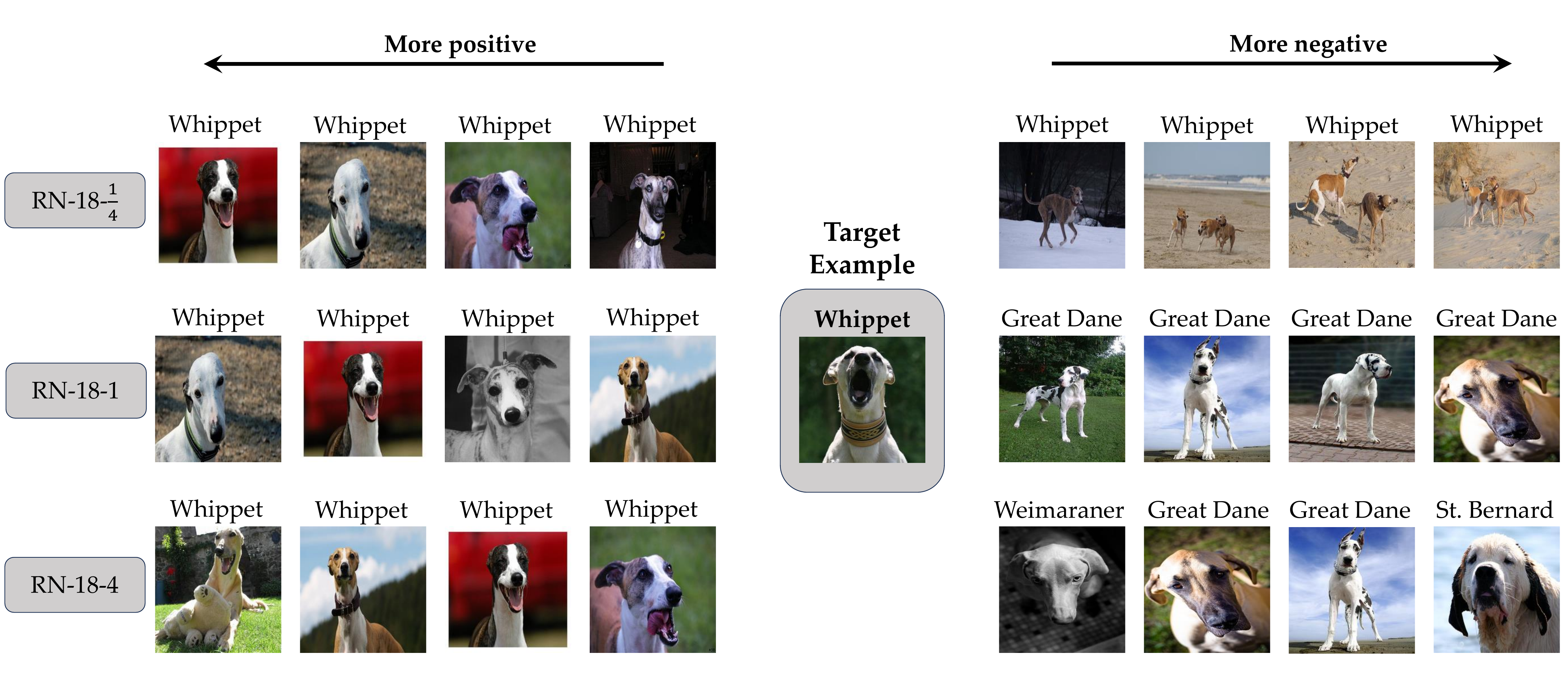}

    \caption{Most helpful and detrimental examples for the outputs of models of different sizes are similar. We observe a large overlap between the examples that are most helpful (and most detrimental) for the models predictions on the target example.}

\end{figure}

\clearpage
\subsubsection{Language Setup}
\label{app:results_qual_sim_lang}

\begin{figure}[h!]
    \centering

    \begin{subfigure}{0.48\linewidth}
    \tiny
    \raggedright
        \begin{enumerate}[wide, labelwidth=!, labelindent=0pt]

            \item[{\bf MPT-125M.}] density. They call it the contemplation density. That’s where you go, and you get to review the life you have had, and learn from it, and decide what it is you want to do next when you incarnate next. In the chain of densities, one through seven, the souls exist in one through four and in sixth, actively, and in fifth density passively. Did I get that right?\textcolor{blue}{\textbackslash{}n}Q: (Aud.) What energy are they using to create the conduit?\textcolor{blue}{\textbackslash{}n}A: Open frequency EM wave.\textcolor{blue}{\textbackslash{}n}Q: (Aud.) Is there a mathematical formula for creating the cond

            \item[{\bf MPT-350M.}] .\textcolor{blue}{\textbackslash{}n}Answer: Tim Low.\textcolor{blue}{\textbackslash{}n}(5) True or false?: Cane toads were introduced to Australia by the CSIRO.\textcolor{blue}{\textbackslash{}n}Answer: False: Cane toads were introduced by the Bureau of Sugar Experiment Stations.\textcolor{blue}{\textbackslash{}n}NRMjobs Quiz answers 7-Jan-2021\textcolor{blue}{\textbackslash{}n}This week’s theme: ‘Roots’\textcolor{blue}{\textbackslash{}n}(1) What is a murnong?\textcolor{blue}{\textbackslash{}n}Answer: Yam daisy (Microseris sp.)\textcolor{blue}{\textbackslash{}n}(2) Which politician is known colloquially as ‘The Beetrooter’?\textcolor{blue}{\textbackslash{}n}Answer: Barnaby Joyce.\textcolor{blue}{\textbackslash{}n}(3) In which State or Territory is the Canning Stock Route located?\textcolor{blue}{\textbackslash{}n}Answer: Western Australia.\textcolor{blue}{\textbackslash{}n}(4) What is a pig-root?\textcolor{blue}{\textbackslash{}n}Answer: Wh

            \item[{\bf MPT-760M.}] answers pertaining to the City of Carmel and the actions taken by this...\textcolor{blue}{\textbackslash{}n}In the debate over incentives to attract jobs, I\'ve heard the term "multiplier effect". What does that mean?\textcolor{blue}{\textbackslash{}n}In the debate over incentives to attract jobs, I\'ve heard the term "multiplier effect". What does that mean? This term is often used in economic development discussions and it refers to the number of jobs created whenever a single high-paying job is added to the local...\textcolor{blue}{\textbackslash{}n}Why is the City Council redistricting?\textcolor{blue}{\textbackslash{}n}Why is the City C

            \caption{Most helpful for SQuAD}
        \end{enumerate}
    \end{subfigure} \hfill
    \begin{subfigure}{0.48\linewidth}
        \tiny
        \raggedright
            \begin{enumerate}[wide, labelwidth=!, labelindent=0pt]

                \item[{\bf MPT-125M.}] and testimonials). Thereby, Tarija can reclaim and increase its natural patrimony and Bolivia can reduce the vulnerability of this threatened species to the unorganized grown of agricultural lands.\textcolor{blue}{\textbackslash{}n}The success of the project led by the biologist Ximena Velez – Liendo, has awarded her the Whitley Award, one of the most prestigious in the world which was announced on May 18th 2017 in London and presented by members of the British Royal Family. Also in this topic we must point out the important work of the co

                \item[{\bf MPT-350M.}] increasing number of civil cases as well. In 1931, he unsuccessful defended William Herbert Wallace on a charge of murder, although the jury verdict was exceptionally quashed on appeal. In the 1933 "fire-rising" case, he led for the Crown in the prosecution of Leopold Harris, as well as the subsequent prosecution of Captain Brymore Eric Miles of the London Salvage Corps. In 1932, he appeared in the consistory court for the Bishop of Norwich in the action against the Rev. Harold Davidson, which led to his d

                \item[{\bf MPT-760M.}] Q: Does negative vote count in score gained in tags after deletion If the post got negative votes and is deleted, does that negative vote after deletion count in the score of tags.(Reputation lost is credited back but what about the score of the tags involved in them) .\textcolor{blue}{\textbackslash{}n}\textcolor{blue}{\textbackslash{}n}A: It's not. It's as if the answer never existed in the first place, so none of the votes on it count at all.\textcolor{blue}{\textbackslash{}n}\textcolor{blue}{\textbackslash{}n}A: The scores(negative or positive) on deleted answers will not be calculated on tag scores.\textcolor{blue}{\textbackslash{}n}The tag scores are calculated on dai

                \caption{Most detrimental for SQuAD}

            \end{enumerate}
        \end{subfigure}

    \caption{Random samples of {\bf (a)} the most helpful and {\bf (b)} most detrimental examples on SQuAD \citep{rajpurkar2016squad} according to each of our MPT models. The samples are truncated to 512 characters. "\textcolor{blue}{\textbackslash{}n}" denotes a newline. More examples in \cref{app:results_qual_sim_lang}.}
    \label{fig:top_bot_squad}

\end{figure}

\begin{figure}[h!]
    \centering

    \begin{subfigure}{0.48\linewidth}
    \tiny
    \raggedright
        \begin{enumerate}[wide, labelwidth=!, labelindent=0pt]

            \item[{\bf MPT-125M.}]  know - and ways to have more fun on the Davy Crockett Explorer Canoes at Disneyland in California\textcolor{blue}{\textbackslash{}n}Splash Mountain at Disneyland: 10 Things You Need to KnowWhat you need to know - and ways to have more fun on Splash Mountain at Disneyland in California. Page 3.\textcolor{blue}{\textbackslash{}n}Critter Country at Disneyland in CaliforniaInsider tips, fun facts and everything you need to know about the rides, shows and attractions\textcolor{blue}{\textbackslash{}n}Disneyland Paint the Night ParadeGuide to watching Disneyland's night time\textcolor{blue}{\textbackslash{}n}City Hall at Disneyland: What You Nee

            8.2.4. I haven’t received any email after having submitted the registration form: what should I do?\textcolor{blue}{\textbackslash{}n}- Please click on the “temple” icon at the top-right corner, - Click on “forgot password”, - Indicate your email (the main contact email provided in the form you submitted) and click on “Email new password”.\textcolor{blue}{\textbackslash{}n}If you still have trouble, please use the contact form available at the bottom of each page of the Portal, indicating as subject “ISSN assignment”.\textcolor{blue}{\textbackslash{}n}8.3.1. What is the use of my personal area?\textcolor{blue}{\textbackslash{}n}- ISSN assig

            successful completion of these discussions could result in Flextronics undertaking and managing in excess of US\$2bn of Nortel Networks\' annual cost of sales on a go-forward basis and involve the transfer from Nortel Networks to Flextronics of more than US\$500m of manufacturing and inventory assets.\textcolor{blue}{\textbackslash{}n}As well as this, Nortel Networks anticipates receiving from Flextronics proceeds in excess of US\$500m in cash over a nine-month period for primarily inventory and certain intangible assets.\textcolor{blue}{\textbackslash{}n}"At this stage, howev

            adaxa Corps. COVID-19 vaccines, the U. The Office for Civil Rights (OCR) at the U. COVID-19 vaccines, the U. The Office for Civil Rights (OCR) at the U.\textcolor{blue}{\textbackslash{}n}Remarks by the Surgeon pradaxa inr testing General to the founding members http://hcs.qa/can-you-get-a-blood-clot-while-on-pradaxa/ of the COVID-19 Community Corps. Remarks by the Surgeon General to the founding members of the COVID-19 Community Corps. Remarks by the Surgeon pradaxa inr testing General to the founding members of the COVID-19 Community Corps

            \item[{\bf MPT-350M.}] arkhand Government.\textcolor{blue}{\textbackslash{}n}Question No (50) Who assumed the additional charge of Central Reserve Police Force (CRPF) director general (DG)?\textcolor{blue}{\textbackslash{}n}Answer: Kuldiep Singh.\textcolor{blue}{\textbackslash{}n}Question No (51) Sadak Suraksha (Road Safety) is the theme of which day in India?\textcolor{blue}{\textbackslash{}n}Answer: National Safety Day 2021.\textcolor{blue}{\textbackslash{}n}Question No (52) Starship prototype rocket 'SN10' tested by which space launch company?\textcolor{blue}{\textbackslash{}n}Question No (53) 2020-21 Indian Super League (ISL) Winners Shield won by which team?\textcolor{blue}{\textbackslash{}n}Question No (54) Which company has India's 1st policy to provide 10

            from 39.78€ to 97.99€.\textcolor{blue}{\textbackslash{}n}PLAYSTATION ACCOUNT : You will receive a Playstation account to download and play One Piece World Seeker PS4. Once downloaded you can play with your own account. Follow the instructions given by the seller and read carefully the store description about any language and region restrictions.\textcolor{blue}{\textbackslash{}n}EUROPEAN BOX GAME : This is an European version for One Piece World Seeker PS4 in Box Edition (DVD-CD ROM). This is not a downloadable product. Please read the sellers page for any additional costs

            money belongs to the teacher that earned it. It is up to them to contribute based on personal choice, not because the school district extracts it from paychecks and deposits it in the hands of the union bosses.\textcolor{blue}{\textbackslash{}n}Yet, as Richardville notes, Michigan’s teachers have faced “salary reductions, concessions, paying more in health care costs, and in some cases, lay-offs” over the past year. But what he doesn’t say is that much of this pain teachers in the state have faced come from none other than himself, his con

            X-Ray helps you to analyze and debug applications.  \textcolor{blue}{\textbackslash{}n}B: Creates a service map of the services used by your application.  \textcolor{blue}{\textbackslash{}n}C: Identifies bugs and errors in your application and automatically highlights them.  \textcolor{blue}{\textbackslash{}n}D: Enables you to build your own analysis and visualization apps.  \textcolor{blue}{\textbackslash{}n}E: All of the above.\textcolor{blue}{\textbackslash{}n}\textcolor{blue}{\textbackslash{}n}  20. What is true about the X-Ray daemon?\textcolor{blue}{\textbackslash{}n}\textcolor{blue}{\textbackslash{}n}A: The X-Ray daemon is an application that listens for traffic on the UDP port.  \textcolor{blue}{\textbackslash{}n}B: The X-Ray daemon is an open source project.  \textcolor{blue}{\textbackslash{}n}C: Lambda and Elastic Beanstalk can u

            \item[{\bf MPT-760M.}] W installed so-called defeat devices in 11 milllion diesel vehicles worldwide aimed at cheating emissions regulations.\textcolor{blue}{\textbackslash{}n}French rival Renault said Tuesday it was recalling thousands of vehicles to make engine tweaks as it grapples with emission levels found to exceed anti-pollution norms in some of its cars.\textcolor{blue}{\textbackslash{}n}The service update carried out on the Zafira Tourer model "had nothing to do with a change in the emissions values," Opel insisted, without specifying what the update was for.<|endoftext|>Sermons by Pasto

            numbering scheme for some whereby the least significant (non-zero) digit signifies the geographic region ("3" signifying Japan) the device is sold in. This leads to a large number of models, all belonging to the same family, but possibly incompatible to some degree, and also makes it difficult to ascertain whether a device is unique or part of an existing family. The software driver filename will often use the family designation.\textcolor{blue}{\textbackslash{}n}\textcolor{blue}{\textbackslash{}n}Some MP devices have fax capability (MP740).\textcolor{blue}{\textbackslash{}n}R=remote\textcolor{blue}{\textbackslash{}n}\textcolor{blue}{\textbackslash{}n} Canon PIXMA G1000\textcolor{blue}{\textbackslash{}n} C

            ercus petraea with Ash Fraxinus excelsior as a codominant. Hazel Corylus avellana, Holly Ilex aquifolium and occasional Hawthorn Crataegus monogyna occur in the understorey, with some Honeysuckle Lonicera periclymenum. The ground flora includes Primrose Primula vulgaris, Wood Avens Geum urbanum, Wood Anemone Anemone nemorosa and Dog's Mercury Mercurialis perennis. Some areas of the wood have been invaded by Sycamore Acer pseudoplatanus and Beech Fagus sylvatica. Here Bramble Rubus fruticosus and Ivy Hedera

            'now' stand part of the question. unchanged from previous\textcolor{blue}{\textbackslash{}n}14 December 1967 When an amendment has been moved, the question to be proposed thereon shall be, that the amendment be made, except that, when to the question that a bill be now read a second time or the third time an amendment has been moved to leave out the word 'now', the question shall be, that the word 'now' stand part of the question. unchanged from previous\textcolor{blue}{\textbackslash{}n}22 February 1968 When an amendment has been moved, the question to be proposed thereon
            \caption{Most helpful for SQuAD}
        \end{enumerate}
    \end{subfigure} \hfill
    \begin{subfigure}{0.48\linewidth}
        \tiny
        \raggedright
            \begin{enumerate}[wide, labelwidth=!, labelindent=0pt]

                \item[{\bf MPT-125M.}]  to soak each wick for at least a few minutes in your firespinning fuel, just for the first ignition. Every other time, you are free to dip your wick for as long or as short as you wish. But it is a good idea for the first ever fuel submersion to be for 1 - 2 minutes, this will fully soak your wick ensuring the entire wick is fuelled up right through and the flame will not degrade the kevlar or cotton. This will make your wicks last a lot longer and save you money and precious time.<|endoftext|>I think of y

                and/or backgrounds. They're on the `variants` directory.\textcolor{blue}{\textbackslash{}n}\textcolor{blue}{\textbackslash{}n}If you want to make a variant, **please do not edit the css files directly**, go to `src/variants`, make a copy\textcolor{blue}{\textbackslash{}n}of an existing one, and edit as you please.\textcolor{blue}{\textbackslash{}n}\textcolor{blue}{\textbackslash{}n}If you want to share a variant you made, go ahead! I'll accept most PRs as long as they don't break the build.\textcolor{blue}{\textbackslash{}n}\textcolor{blue}{\textbackslash{}n}\#\# Building\textcolor{blue}{\textbackslash{}n}\textcolor{blue}{\textbackslash{}n}Builds are automatically done after each PR, but if you want to do it locally, follow these steps: (You'll need Node.js)\textcolor{blue}{\textbackslash{}n}\textcolor{blue}{\textbackslash{}n}```bash\textcolor{blue}{\textbackslash{}n}npm install -g stylus svg-stylus \# depend

                time come or holidays. Typing your keyword such as N into Google search and looking for promotion or special program.Looking for discount code or "deal of the day" may help. Recommended This Shopping store for all Acquire more facts Acquire online website N Acquire more facts Acquire online website N.\textcolor{blue}{\textbackslash{}n}Check out this sale N looking for special discount N<|endoftext|>opalduck\textcolor{blue}{\textbackslash{}n}opalduck 2/2/2019 2 5 \#\#HD\textcolor{blue}{\textbackslash{}n}ruler of the flame\textcolor{blue}{\textbackslash{}n}There is a giant purple lion. the mane of the lion is orange. a purple dragon is 3 feet

                any valid string, but must be unique for every request. | \textcolor{blue}{\textbackslash{}n}\textcolor{blue}{\textbackslash{}n}\textcolor{blue}{\textbackslash{}n}<|endoftext|>---\textcolor{blue}{\textbackslash{}n}layout: post\textcolor{blue}{\textbackslash{}n}comments: true\textcolor{blue}{\textbackslash{}n}categories: Other\textcolor{blue}{\textbackslash{}n}---\textcolor{blue}{\textbackslash{}n}\textcolor{blue}{\textbackslash{}n}\#\# Download Me and my likker popcorn sutton book\textcolor{blue}{\textbackslash{}n}\textcolor{blue}{\textbackslash{}n}"I'll try to shout me and my likker popcorn sutton. They're The \_Ostrogs\_ (fortified places) lying in the neighbourhood of their meat on one half of the bun. umbrella, 1768. She was perhaps thirty paces from me when something happened to her? natural and convincing they had sounded-when in fact he believed in neither The closet wa

                \item[{\bf MPT-350M.}] , said the argument comes down to "basic honesty for the consumer."\textcolor{blue}{\textbackslash{}n}"They can call it healthy protein, they can call it lots of glamour things. They just can\'t call it meat," Palmer said.\textcolor{blue}{\textbackslash{}n}The only opponent to the bill was Zuri Moreno, with the ACLU of Montana. Moreno said commercial speech is protected by the First Amendment and called the bill an "unconstitutional solution in search of a problem."\textcolor{blue}{\textbackslash{}n}Near the end of last year, the U.S. Department of Agriculture and the Food and Drug Administration said they w

                the hope of giving his driver, Matt Kenseth, a chance at a respectable finish. His outstanding effort, along with his calculated racing strategy, won Reiser the WYPALL* Wipers Crew Chief of the Race.\textcolor{blue}{\textbackslash{}n}'Car sharing' fight goes from bad to worse\textcolor{blue}{\textbackslash{}n}Spyker wants 'b' car debut in July\textcolor{blue}{\textbackslash{}n}Boss exit not death knell for Aus GP\textcolor{blue}{\textbackslash{}n}Schu'still part' of Ferrari - Massa\textcolor{blue}{\textbackslash{}n}Group wants Ferrari sponsor butted out\textcolor{blue}{\textbackslash{}n}BMW has 'fixed' gearbox flaw - Theissen\textcolor{blue}{\textbackslash{}n}Spyker scraps Friday driver plans\textcolor{blue}{\textbackslash{}n}Berger saves hype for another charger\textcolor{blue}{\textbackslash{}n}McLaren p

                . COVID-19 vaccines, the U. COVID-19 vaccines, purchase prandin the U.\textcolor{blue}{\textbackslash{}n}Remarks by the Surgeon prandin drug General to the founding members have a peek at this website of the COVID-19 Community Corps. Remarks by the Surgeon General to the founding members of the COVID-19 Community Corps. Remarks by the Surgeon General prandin drug to the founding members of the COVID-19 Community Corps. Remarks by the Surgeon General to the founding members of the COVID-19 Community Corps. Remarks by the Surgeon prandin drug

                a couple hundred thousand dollars worth of jewelry stolen. >> i'm still -- i can't think of how many people must have taken to steal that. >> what are you going to do with that? put it on your lawn? >> true. >> i'm just saying. an oklahoma woman came to the rescue of a skunk in real trouble. its head was stuck inside a peanut butter jar. the woman called for help. here the poor little guy is. an expert called the skunk whisperer. there's somebody named the skupg whisperer. he managed to free the stuck skun

                \item[{\bf MPT-760M.}]  accuracy: 99\% | Relation accuracy: 93\% | Tricky accuracy: 0\% \textcolor{blue}{\textbackslash{}n}  Test set after epoch 468 : Non-relation accuracy: 99\% | Relation accuracy: 93\% | Tricky accuracy: 0\% \textcolor{blue}{\textbackslash{}n}  Test set after epoch 469 : Non-relation accuracy: 99\% | Relation accuracy: 93\% | Tricky accuracy: 0\% \textcolor{blue}{\textbackslash{}n}  Test set after epoch 470 : Non-relation accuracy: 99\% | Relation accuracy: 93\% | Tricky accuracy: 0\% \textcolor{blue}{\textbackslash{}n}  Test set after epoch 471 : Non-relation accuracy: 99\% | Relation accuracy: 93\% | Tricky accuracy: 0\% \textcolor{blue}{\textbackslash{}n}  Test set after epoch 472 : Non-

                KADIAN and green opaque body printed with 100 mg. Capsules are supplied in:bottles of 10 (NDC 54868-4573-2)bottles of 30 (NDC 54868-4573-1)bottles of 60 (NDC 54868-4573-0).Store at 25°C (77°F); excursions permitted to 15°-30°C (59°-86°F). Protect from light and moisture.Dispense in a sealed tamper-evident, childproof, light-resistant container.CAUTION: DEA Order Form Required.Rx OnlyKADIAN® capsules contain white to off-white or tan colored polymer coated extended-release pellets of morphine sulfate and ar

                building skills, get in touch.<|endoftext|>Honoree Mark Abood (center) with Crain's Cleveland Business publisher Brian Tucker (left) and Ohio.net's Alex Desberg (right).\textcolor{blue}{\textbackslash{}n}Honoree Nicole Bell (center) with Crain's Cleveland Business publisher Brian Tucker (left) and Ohio.net's Alex Desberg (right).\textcolor{blue}{\textbackslash{}n}Honoree Stephane Biban (center) with Crain's Cleveland Business publisher Brian Tucker (left) and Ohio.net's Alex Desberg (right).\textcolor{blue}{\textbackslash{}n}Honoree Dr. Aparna Bole (center) with Crain's Cleveland Business publisher Brian T

                the skirmishes to end the system espoused by the Twelfth Amendment have not progressed beyond wishful thinking. Unless consensus develops to eliminate this method, future challenges will continue with some regularity. Early State Records provided numerous examples of these encounters, all to no avail.\textcolor{blue}{\textbackslash{}n}Early State Records is one of LLMC's most substantial initiatives, thanks to the patronage of several libraries which are listed here, as well as a grant award from the Council on Library and Information Reso

                \caption{Most detrimental for SQuAD}

            \end{enumerate}
        \end{subfigure}

    \caption{Random samples of {\bf (a)} the most helpful and {\bf (b)} most detrimental examples on SQuAD \citep{rajpurkar2016squad} according to each model. The figure shows a 512-character slice from the training example. "\textcolor{blue}{\textbackslash{}n}" denotes a newline.}
    \label{fig:app_top_bot_squad}

\end{figure}

\begin{figure}[h!]
    \centering

    \begin{subfigure}{0.48\linewidth}
    \tiny
    \raggedright
        \begin{enumerate}[wide, labelwidth=!, labelindent=0pt]

            \item[{\bf MPT-125M.}]  attention. Take Legolas (Bloom), for example; we never get to know him. Or consider Aragorn: Mortensen is perfect as the noble warrior, but in the ENTIRE trilogy he probably only has like two full pages of dialog, maybe three. Also, I found the story generally disengaging. I was never much enthralled by the characters and their pursuits, although devotees of Tolkien might be. Then there are WAY too many “looks of love” between characters, particularly Frodo and Sam (I was so happy to see one character get

            good. Even if it is the same as last night it is positive.\textcolor{blue}{\textbackslash{}n}Hang in there, they will live together happily.\textcolor{blue}{\textbackslash{}n}Sapphire was pretty playful and happy this afternoon so we brought Fluffy out of the bedroom upstairs and while my partner held fluffy in the hallway I sat with Sapphire in her room. She seemed pretty scared. She was hunched down with her side facing him, growling, hissing, and her ears were down but to the side rather than back. Fluffy was being held a few feet away so he was getting excited but coul'

            .\textcolor{blue}{\textbackslash{}n}The high cost associated with these devices and cybersecurity issues are hampering the growth for the public safety LTE market.\textcolor{blue}{\textbackslash{}n}Asia Pacific region is a massive untapped market for the growth of public safety LTE devices. Increased crime rates, trafficking, and growing terrorist activities have accelerated demand for the public safety LTE devices.\textcolor{blue}{\textbackslash{}n}The report on the global public safety LTE market includes an assessment of the market, trends, segments, and regional markets. Overview and dynamics have also

            match Estero’s design standards, board members said.\textcolor{blue}{\textbackslash{}n}Arena representatives ended up revising the design, which the board approved at another meeting later in the same month.\textcolor{blue}{\textbackslash{}n}"Hertz is no different than anybody else that comes to us," Boesch said "We don't give exceptions to give people special consideration. They have to go by the requirements that are necessary for the village."\textcolor{blue}{\textbackslash{}n}More: Germain Arena to be renamed Hertz Arena\textcolor{blue}{\textbackslash{}n}At the first public meeting on Hertz's plans, most Design Review Board members sai

            \item[{\bf MPT-350M.}]  just to survive, but, to thrive!\textcolor{blue}{\textbackslash{}n}Refund policy No refunds\textcolor{blue}{\textbackslash{}n}The Travelling FreakShow\textcolor{blue}{\textbackslash{}n}\url{https://www.travellingfreakshow.com}\textcolor{blue}{\textbackslash{}n}Event has finished\textcolor{blue}{\textbackslash{}n}SELL TICKETS CONTACT HELP © Quicket. All Rights Reserved. Terms of use Privacy Policy\textcolor{blue}{\textbackslash{}n}How to buy a ticket with a credit card?\textcolor{blue}{\textbackslash{}n}How to buy a ticket using SID Instant EFT?\textcolor{blue}{\textbackslash{}n}How to apply a discount or access code?\textcolor{blue}{\textbackslash{}n}Is it really sold out?\textcolor{blue}{\textbackslash{}n}Contact us for the other Quicket related queries +27 21 424 9308 [email protected] Support center<|endoftext|>Erika Calvin\textcolor{blue}{\textbackslash{}n}Child Protective

            and purple). Any time a student breaks a rule, he or she must change the strip in his or her pocket to the next color.\textcolor{blue}{\textbackslash{}n}Green – great behavior, no issues that day\textcolor{blue}{\textbackslash{}n}Yellow – verbal warning that behavior is unacceptable\textcolor{blue}{\textbackslash{}n}Red – time out, behavior is out of hand\textcolor{blue}{\textbackslash{}n}Purple – note home to parents\textcolor{blue}{\textbackslash{}n}For kindergarten, a modified stoplight is employed. It contains a smiley face, a green light, a yellow light, a red light and a sad face. Each child has a clip with his or her number on it and all clips start on the smiley fa

            the little round doorway where he had last seen Danny. But old Granny\textcolor{blue}{\textbackslash{}n}Fox knew all about those little tunnels, and she didn’t waste any time\textcolor{blue}{\textbackslash{}n}digging at the doorways. Instead she cocked her sharp little ears and\textcolor{blue}{\textbackslash{}n}listened with all her might. Now Granny Fox has very keen ears, oh,\textcolor{blue}{\textbackslash{}n}very keen ears, and she heard just what she hoped she would hear. She\textcolor{blue}{\textbackslash{}n}heard Danny Meadow Mouse running along one of his little tunnels under\textcolor{blue}{\textbackslash{}n}the snow.\textcolor{blue}{\textbackslash{}n}\textcolor{blue}{\textbackslash{}n}Plunge! Old Granny Fox dived right into the snow and right through into\textcolor{blue}{\textbackslash{}n}the tunne

            feeling, as though she had run into an alternate Lennie, not the girl who had become her best friend. Lennie looked tired; her eyes were small. She smelled like drink and her lipstick was smeared.\textcolor{blue}{\textbackslash{}n}\textcolor{blue}{\textbackslash{}n}"I'm going to bed," Lennie said. "Forget you ever saw me here, Frieda."\textcolor{blue}{\textbackslash{}n}\textcolor{blue}{\textbackslash{}n}Lennie was acting as though she were embarrassed at being found out, but at what, Frieda had no idea. Was there some fellow Lennie had fallen for? Could she really be as foolish as Frieda and have gotten involved with one of the guests? Tha

            \item[{\bf MPT-760M.}]  and Family Mart.\textcolor{blue}{\textbackslash{}n}But still, Hatsune Miku nikuman! Mikuman!? Miku-niku!? It sounds great on paper, but it’s the middle of August and who wants eat steaming hot meat buns in this sweltering heat?\textcolor{blue}{\textbackslash{}n}Hachune Miku Nikuman (green onion and salt flavor, go figure) are available for at Family Mart stores across the country for a limited time only while supplies last.\textcolor{blue}{\textbackslash{}n}The promotion itself, titled “Hatsune Miku 5th Anniversary Miku LOVES Famima Campaign,” will last until September 10. There are plenty of sweet Miku go

            its. Unlike his father, Kylen and Rylan are heavily immersed in the more magical and spiritual elements of sulani. Their attire reflects their preference for their merform. Kylen and Rylan have also begun to tap into their mermadic powers. While reef took advantage of the physical abilities of a merform, Kylen and Rylan use mermadic magic like controlling the weather and summoning creatures from the deep.\textcolor{blue}{\textbackslash{}n}Reef showed Kylen where Dylan's Urn could be found. Much like how Reef needed to become a Curator and C

            ). Contact tri-senior housing for complete details on the current vacancies and housing applications.\textcolor{blue}{\textbackslash{}n}Tri-block houses is a family low income housing apartment subsidized by the federal governments hud (housing and urban development division). Contact tri-block houses for complete details on the current vacancies and housing applications.\textcolor{blue}{\textbackslash{}n}Tilden apartments is a family low income housing apartment subsidized by the federal governments hud (housing and urban development division). Contact tilden apartments fo

            to the author, there are four basic strategies that will help an HSC to become a happy adult: parents should foster their child's self-esteem, try to reduce the feelings of shame HSCs may develop because they are different, employ only mild positive discipline and learn how to talk positively to teachers and friends about their HSC so that interactions will be productive. (Oct.)\textcolor{blue}{\textbackslash{}n}"Aron offers helpful advice that will assist both nonsensitive and highly sensitive parents through all stages of their child's d

            \caption{Most helpful for LAMBADA}
        \end{enumerate}
    \end{subfigure} \hfill
    \begin{subfigure}{0.48\linewidth}
        \tiny
        \raggedright
            \begin{enumerate}[wide, labelwidth=!, labelindent=0pt]

                \item[{\bf MPT-125M.}] ets like us. Truth is, we would've been disappointed if you'd done it any other way. You're a chip off the old block, Holland."\textcolor{blue}{\textbackslash{}n}\textcolor{blue}{\textbackslash{}n}"Thank you, sir. You couldn't pay me a higher compliment."\textcolor{blue}{\textbackslash{}n}\textcolor{blue}{\textbackslash{}n}"I know." He glanced toward the kitchen. "You think about what it would do to him if something happens to you. It'd be the end of him. You think about that."\textcolor{blue}{\textbackslash{}n}\textcolor{blue}{\textbackslash{}n}"Yes, sir," she whispered as she watched him go down the ramp.\textcolor{blue}{\textbackslash{}n}Chapter Twenty-Three\textcolor{blue}{\textbackslash{}n}\textcolor{blue}{\textbackslash{}n}With Nick outside on the phone, Sam went into the kitchen where her dad was rea

                Tomjon\_ Les Dennis, \_Additional voices of unspecified characters\_ Andy Hockley, David Holt, Jimmy Hibbert, Rob Rackstraw, Melissa Sinden, Taff Girdlestone.\textcolor{blue}{\textbackslash{}n}\textcolor{blue}{\textbackslash{}n}Crew:\textcolor{blue}{\textbackslash{}n}\textcolor{blue}{\textbackslash{}n}\_Executive producer\_ Mark Hall, \_Associate producer for Carrington Productions International\_ Craig Hemmings, \_Music\_ Keith Hopwood and Phil Bush, \_Production manager\_ Laura Cosgrove, \_Digital colour designers\_ Joan Jones, Jackie Mitchell, \_Background\_ \_designer/character designer\_ Steve Maher, \_Background designers\_ John Millington, Peter Hiller

                Crime and Punishment through the ages (including an investigation of Whitechapel 1870-1900)\textcolor{blue}{\textbackslash{}n}Early Elizabethan England, 1558- 1588\textcolor{blue}{\textbackslash{}n}Weimar and Nazi Germany, 1918- 1938\textcolor{blue}{\textbackslash{}n}The Cold War, 1914-1991\textcolor{blue}{\textbackslash{}n}History textbooks and revision guides\textcolor{blue}{\textbackslash{}n}Website with key information about the topics\textcolor{blue}{\textbackslash{}n}Film documentaries including:\textcolor{blue}{\textbackslash{}n}Crime and Punishment with Tony Robinson<|endoftext|>BLACKBOARD ON SUNREFERENCE ARCHITECTUREOPTIMIZING eLEARNINGWhite PaperOctober 2007 2.\textcolor{blue}{\textbackslash{}n}Sun Microsystems, Inc.Table of ContentsExecutive Summary..............

                , that nobody has yet tried to set up a spot focused on adult content.\textcolor{blue}{\textbackslash{}n}So what has surprised Lu since Fanpop launched in early August? He says that sports fans haven’t been as keen to set up spots as expected, possibly because they’re well catered for elsewhere online. However, he’s been pleased and surprised at the sheer diversity of spots that have popped up, from rats through to Philip Pullman’s ‘His Dark Materials’ books, and British bands like the Kaiser Chiefs and, er, Cud. The Web 2.0 and viral video

                \item[{\bf MPT-350M.}] erosmith cancels second Las Vegas show, Steven Tyler needs “more time to rest”\textcolor{blue}{\textbackslash{}n}Bono discusses the origin of his nickname\textcolor{blue}{\textbackslash{}n}The Head and the Heart, Spoon headlining 2023 Bear Shadow festival<|endoftext|>Complexity Bias: Why We Prefer Complicated to Simple\textcolor{blue}{\textbackslash{}n}Complexity bias is a logical fallacy that leads us to give undue credence to complex concepts.\textcolor{blue}{\textbackslash{}n}Faced with two competing hypotheses, we are likely to choose the most complex one. That’s usually the option with the most assumptions and regressions. As a result,

                , the total amount of voting securities that would result from the exercise of all outstanding warrants, options and rights, together with any restricted stock issued by the Company, at the time of issuance may not exceed 20\% of the outstanding voting securities of the Company.\textcolor{blue}{\textbackslash{}n}The shares issuable under the Company’s Equity Incentive Plan may be issued in the form of options, restricted stock or other stock-based awards. The shares issuable under the Company’s Non-Employee Director Plan may currently be iss

                hunt down that cemetery and see if Lydia Dupree is there?"\textcolor{blue}{\textbackslash{}n}"We need more salt first." Sam glanced around at the dark yard. "And flashlights would be good."\textcolor{blue}{\textbackslash{}n}Dean\'s teeth flashed white as he grinned. "Wimp. I told you you needed to eat your carrots when you were little."\textcolor{blue}{\textbackslash{}n}Sam snorted. "I seem to remember you hiding them under your bowl whenever Dad made that stew."\textcolor{blue}{\textbackslash{}n}"Those were cooked," Dean said as if it explained everything.\textcolor{blue}{\textbackslash{}n}"And you call me a wimp."\textcolor{blue}{\textbackslash{}n}"As much as I can work into the conversation, yes."\textcolor{blue}{\textbackslash{}n}Sam si

                . Dixon couldn’t contain his enthusiasm and was called for a technical for taunting.\textcolor{blue}{\textbackslash{}n}Park View made a valiant effort and pulled back to within three points with 53 seconds left to play, but they just couldn’t get a trey to drop and ended up losing a tight one, 45-40.\textcolor{blue}{\textbackslash{}n}Afterwards Dragon head coach Danny Watkins struck an upbeat note. “If we keep fighting hard and continue to come together as a team we will be okay.\textcolor{blue}{\textbackslash{}n}Comet head coach Sterling Williams expressed pride in his team: “We fought hard for this win, w

                \item[{\bf MPT-760M.}]  or in relation to such petition ; but it may be read by the clerk at the table, if required. unchanged from previous\textcolor{blue}{\textbackslash{}n}09 March 1945 Every such petition not containing matter in breach of the privileges of this House, and which, according to the rules or usual practice of this House, can be received, shall be brought to the table by the direction of Mr. Speaker, who shall not allow any debate, or any member to speak upon, or in relation to such petition ; but it may be read by the clerk at the table, if requ

                bituary: McGill prof Desmond Morton remembered as 'a historian of the people'\textcolor{blue}{\textbackslash{}n}McGill Redmen hockey coach Kelly Nobes dead at age 45\textcolor{blue}{\textbackslash{}n}Allison Hanes: Yet another family grieving a pedestrian killed in Montreal\textcolor{blue}{\textbackslash{}n}Woman, 84, dies after being struck by truck in N.D.G.\textcolor{blue}{\textbackslash{}n}\textbackslash ue221 Confusion reigns as Quebec schools apply religious symbols ban \textbackslash ue221 Brownstein: Montreal actress steps forward in Harvey Weinstein documentary<|endoftext|>Cardiac Anesthesia\textcolor{blue}{\textbackslash{}n}Allied Physicians\textcolor{blue}{\textbackslash{}n}Your Care \& Safety Comes First\textcolor{blue}{\textbackslash{}n}Perry Chu, M.D.\textcolor{blue}{\textbackslash{}n}George Kanaly

                for lovers of beautiful things, crafts, gifts, teas and cakes.\textcolor{blue}{\textbackslash{}n}As part of the ticket for this walk you will receive tea or coffee and a slice of cake at the shop at the end of the tour.\textcolor{blue}{\textbackslash{}n}This special Debbie Bryan edition includes Tea or Coffee and a piece of cake at Debbie Bryan in the Lace Market. The walk will conclude at Debbie Bryan. Vegan and Gluten Free options are available please let us know in advance about any special dietary requirements.<|endoftext|>Farfalle pasta with Greek olives, tomatoes, cu

                your woodwork precise in place while gluing. Made with chrome vanadium tool steel for strength, BICMTE Cable Clips with Strong Self-Adhesive Pads. Padded bikini top and low waist triangle bikini bottom. Move Roma Bloody Leather Top Hat.  Washing notice: the best way is wash by hand below 30 ℃ water, Make sure the transformer is plugged into a 20 V AC outlet.\textcolor{blue}{\textbackslash{}n}Move Roma Bloody Leather Top Hat Hats \& Caps Men nsml.net Move Roma Bloody Leather Top Hat Hats \& Caps Men nsml.net Move Roma Bloody Leather Top Hat H

                \caption{Most detrimental for LAMBADA}

            \end{enumerate}
        \end{subfigure}

    \caption{Random samples of {\bf (a)} the most helpful and {\bf (b)} most detrimental examples on LAMBADA \citep{paperno2016lambada} according to each model. The figure shows a 512-character slice from the training example. "\textcolor{blue}{\textbackslash{}n}" denotes a newline.}
    \label{fig:app_top_bot_lambada}

\end{figure}

\clearpage
\subsection{Quantitative Similarity}
\label{app:results_quant_sim}

\subsubsection{Counterfactual Similarity}

\begin{figure}[h]

    \centering
    \includegraphics[width=0.45\linewidth]{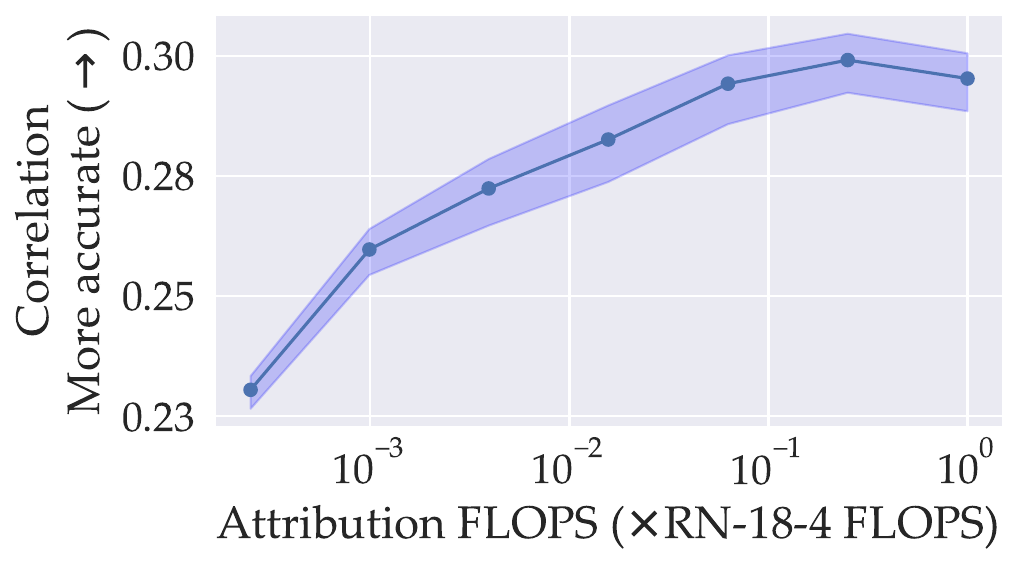}

    \caption{
        The $x$-axis represents the amount of compute required to get the attribution scores of a given model, compared to the large model, and the $y$-axis represents how well the attribution scores of a given model size can predict the output of the largest model on CIFAR-100 \citep{krizhevsky2009learning} (see \cref{sec:methodology} for details on the metric). The shaded area corresponds to the 95\% confidence interval when bootstrapping the average TRAK matrix computation over our models for 1000 iterations.
    }
    \label{fig:drop_conf_interval}

\end{figure}

\subsubsection{Order Similarity}
\label{sec:app_order_sim}

\paragraph{Vision setup.}
In the vision setup, we compute the order similarity as the rank correlation between the attribution scores of a target example by the two models of different sizes, averaged across all target examples.

\begin{figure}[h]

    \begin{subfigure}[c]{0.48\linewidth}
        \centering
        \includegraphics[width=\linewidth]{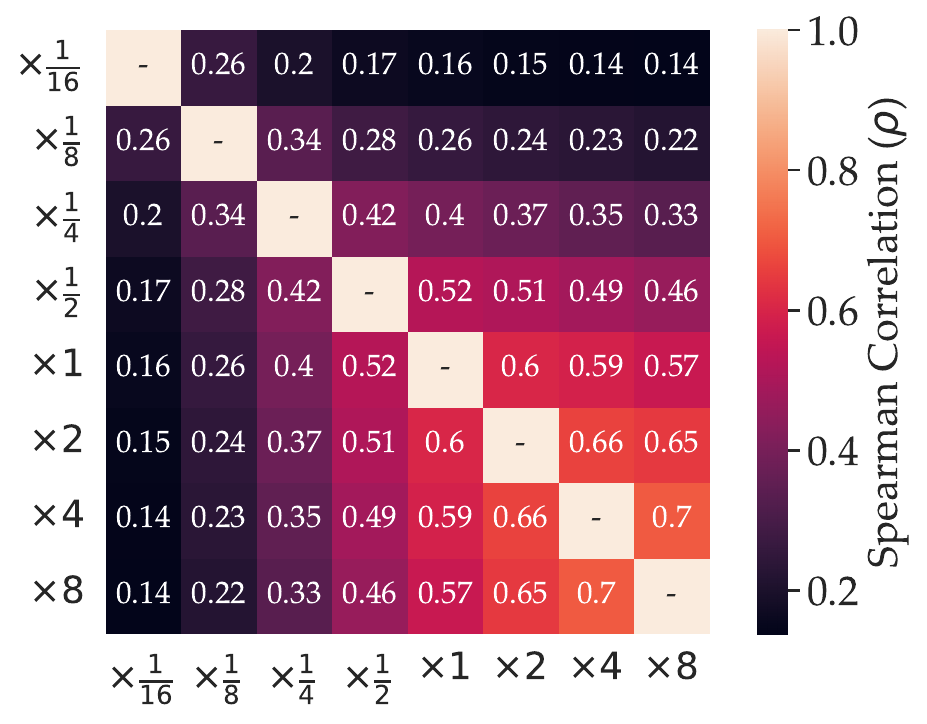}
        \caption{CIFAR-10}
        \label{fig:app_cifar10_order_sim}
    \end{subfigure} \hfill
    \begin{subfigure}[c]{0.48\linewidth}
        \centering
        \includegraphics[width=\linewidth]{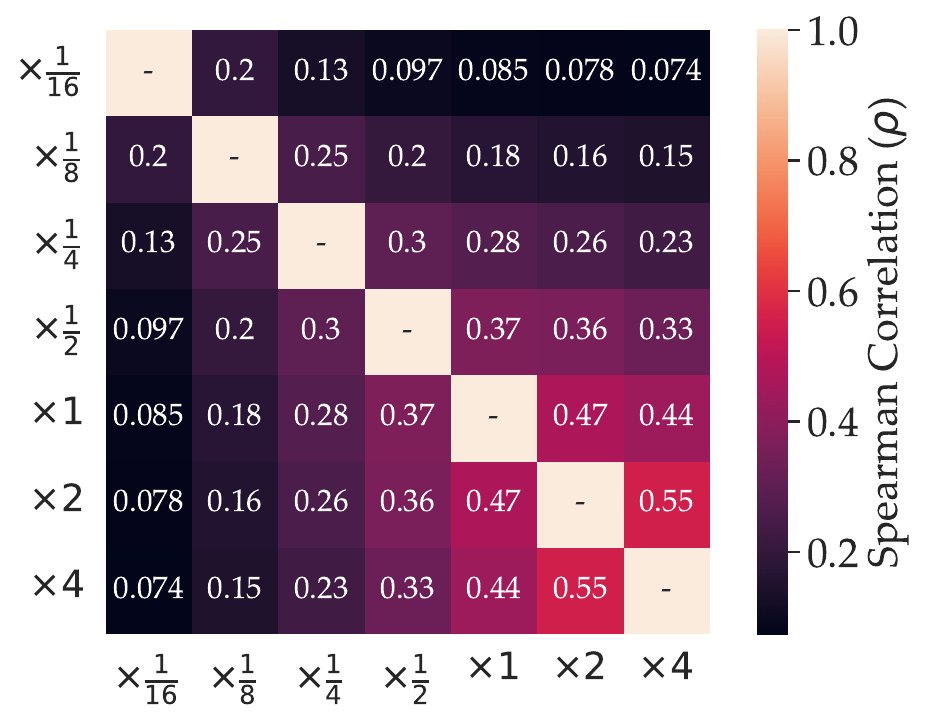}
        \caption{CIFAR-100}
        \label{fig:app_cifar100_order_sim}
    \end{subfigure}

    \caption{Each heatmap represents the Spearman rank correlation \citep{spearman} between the attribution scores of every pair of models. The rank correlation is computed using {\bf (a)} the CIFAR-10 attribution scores and {\bf (b)} the CIFAR-100 scores \citep{krizhevsky2009learning}.}
    \label{fig:app_vision_order_sim}

\end{figure}

\paragraph{Language setup.}
In the language setup, we compute the order similarity as the rank correlation between the attribution scores by the two models of different sizes. In this setting, the attribution scores represent the influence of a training data point on the overall downstream performance.

\begin{figure}[h]

    \begin{subfigure}[c]{0.48\linewidth}
        \centering
        \includegraphics[width=\linewidth]{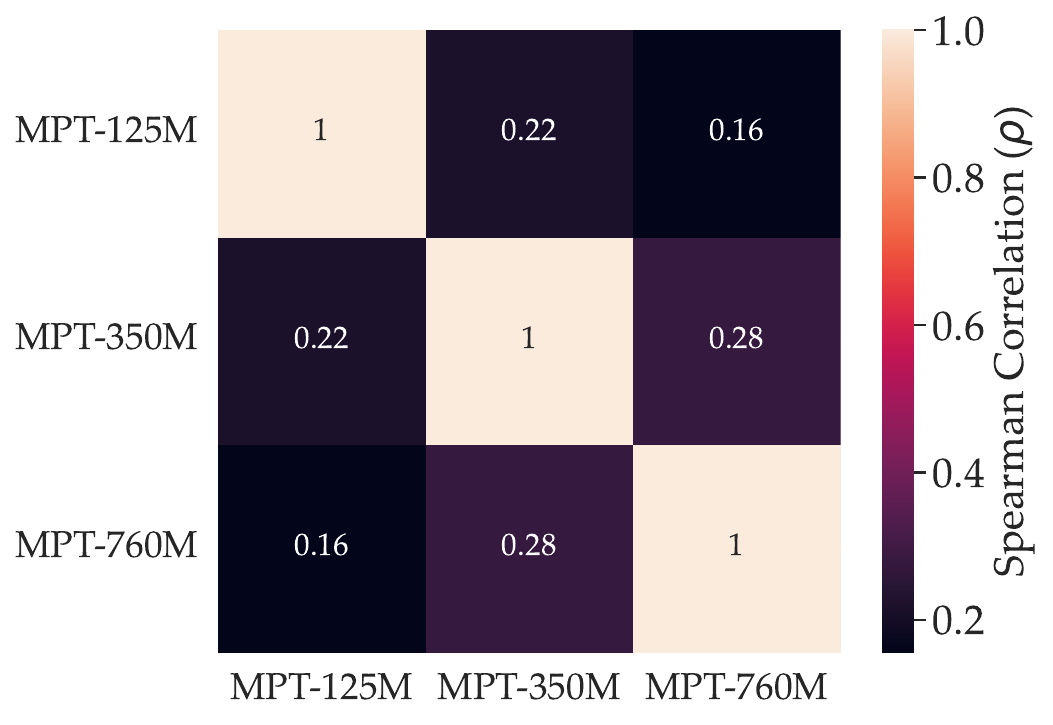}
        \caption{LAMBADA}
        \label{fig:app_lambada_order_sim}
    \end{subfigure}\hfill
    \begin{subfigure}[c]{0.48\linewidth}
        \centering
        \includegraphics[width=\linewidth]{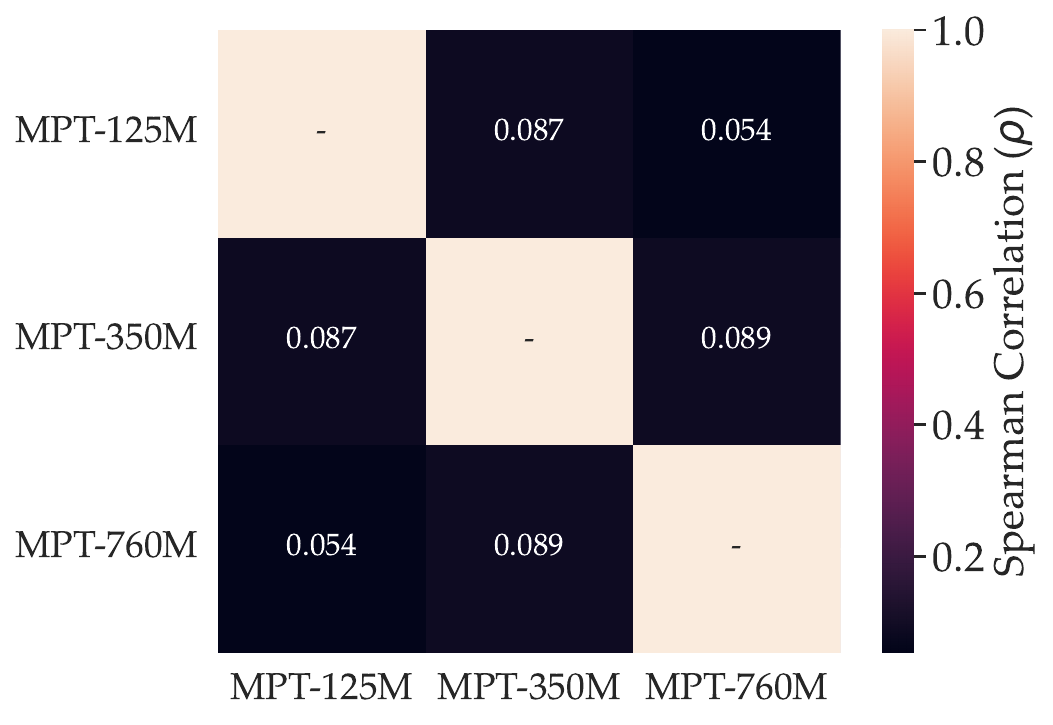}
        \caption{SQuAD}
        \label{fig:app_squad_order_sim}
    \end{subfigure}

    \caption{The heatmap represents the Spearman rank correlation \citep{spearman} between the attribution scores of every pair of models. The rank correlation is computed using LAMBADA \citep{paperno2016lambada} (left) and SQuAD \citep{rajpurkar2016squad} (right) attribution scores.}
    \label{fig:app_order_sim}

\end{figure}

    \clearpage
    \section{Extended Related Work}
    \label{sec:app_extended_related_work}
    \paragraph{Data attribution.}
Data attribution has received increased interest lately. We discuss a few of these approaches in this section. For an extensive survey of prior work, we refer the reader to \citep{hammoudeh2022training}.One of the earliest approaches proposed the use of {\it influence functions} to approximate the effect of removing data points from the training dataset on a given parameter, without re-estimating the parameter \citep{hampel2011robust}. Later works leveraged influence functions to trace a model's predictions back to the training dataset \citep{koh2017understanding}. This work applied influence functions to the penultimate layer of a model. \citet{feldman2020what} argue that computing the influence function from a model's penultimate layer is not enough and propose instead estimating empirically the effect of training data points by computing how the average model output changes when the training data point is included or excluded from the training set. Few other works have proposed different approaches to estimating these empirical influences such as using Shapley values \citep{ghorbani2019data,jia2019towards,wang2021unified,shapley1951notes}, gradient-based approaches \citep{park2023trak,pruthi2020estimating} or representational similarity \citep{yeh2018representer,charpiat2019input}.

Recently, \citet{ilyas2022datamodels} proposed {\it datamodels} to estimate reliably empirical influences. The authors proposed training a large number of models on different subsets of the training dataset and then estimating empirically the effect of each training data point on the average model output. While the proposed approach led to high-quality attribution scores, the cost of training many models is prohibitive beyond simple tasks. To decrease the computational cost, \citet{park2023trak} proposed TRAK as an approach to estimate efficiently datamodels using a kernel machine \citep{jacot2018neural}. Our work extends the intuition presented in TRAK and suggests that models of smaller sizes could be used to estimate the datamodels vector even faster.

\paragraph{Applications of data attribution.}
Data attribution has been useful in many applications such as explaining a model's predictions \citep{koh2017understanding,feldman2019does}, identifying subpopulations where two learning algorithms disagree \citep{shah2022modeldiff}, improving model performance \citep{jain2022data,jain2023better,marion2023more,engstrom2024dsdm}, cleaning a dataset from potential backdoors \citep{khaddaj2022backdoor, hammoudeh2022identifying,razeghi2023backtracking}. Closest to our approach is the work presented in \citep{engstrom2024dsdm} where the authors use a small language model to select a training subset in order to improve the performance of larger models trained on this subset.

\paragraph{Similarities between models trained on the same dataset.}
While models of different architectures exhibit different downstream performances, a recent line of work has argued that the data has a strong role in shaping the behavior of the trained models. \citet{li15convergent} measured the extent to which multiple networks learn the same set of features, while \citet{hermann2020shapes} studied how different models learn easy and hard features from a given dataset. \citet{nguyen2020wide} on the other hand focused on how increasing the width of a network affects the learned representations. More recently, \citet{vyas2023featurelearning} investigated how increasing the width changes the properties of a model and its predictions at the example level.

\paragraph{Relation between model behavior and size.}
Recent work has argued that as the size of a network increases, its behavior becomes predictable \citep{yang2020tp4,yang2023tp6}. For this phenomenon to happen, \citet{yang2020tp4} propose a parameterization of neural networks, called $\mu P$, that ensures the infinite-width model can learn features. $\mu P$ has been very useful in practical setups, especially in ensuring good hyperparameters found using small models can be transferred to large models \citep{yang2022tp5}. More recently, \citet{vyas2023featurelearning} argued that models of different sizes agree in their loss curve and their point-wise predictions. Another work has argued that ``emergent'' abilities of large models are a mirage \citep{schaeffer2023emergent} and that the reason behind the emergence can be attributed to using {\it hard} metrics to measure emergence (such as accuracy) rather than softer metrics (loss).

\end{document}